\documentclass[sigconf]{acmart}
\AtBeginDocument{%
  }

\setcopyright{acmlicensed}
\copyrightyear{2018}
\acmYear{2018}
\acmDOI{XXXXXXX.XXXXXXX}

\acmConference[Conference acronym 'XX]{Make sure to enter the correct
  conference title from your rights confirmation email}{June 03--05,
  2018}{Woodstock, NY}

\acmISBN{978-1-4503-XXXX-X/2018/06}

\usepackage{amsmath}
\usepackage{graphicx}
\usepackage{booktabs}
\usepackage{subcaption} 
\usepackage{multirow}
\usepackage{enumitem}
\usepackage{makecell} 
\usepackage[table]{xcolor}

\usepackage{tcolorbox}
\usepackage{listings}
\usepackage{xcolor}

\usepackage{algorithm}
\usepackage{algorithmic}
\usepackage{amsmath}

\begin{document}

\title{Spatio-Temporal Token Pruning for Efficient High-Resolution GUI Agents}

\author{Zhou Xu}
\affiliation{%
  \institution{Tsinghua University}
  \city{Shenzhen}
  \country{China}}
\email{xu-z25@mails.tsinghua.edu.cn}

\author{Bowen Zhou}
\affiliation{%
  \institution{Tsinghua University}
  \city{Shenzhen}
  \country{China}}
\email{zhoubw25@mails.tsinghua.edu.cn}

\author{Qi Wang}
\affiliation{%
  \institution{Xidian University}
  \city{Xi'an}
  \country{China}}
\email{qiwang0720@stu.xdu.edu.cn}

\author{Shuwen Feng}
\affiliation{%
  \institution{Tsinghua University}
  \city{Shenzhen}
  \country{China}}
\email{fsw25@mails.tsinghua.edu.cn}

\author{Jingyu Xiao}
\authornote{Jingyu Xiao is the corresponding author.}
\affiliation{%
  \institution{The Chinese University of Hong Kong}
  \city{Hong Kong}
  \country{China}}
\email{jyxiao@link.cuhk.edu.hk}

\begin{abstract}
Pure-vision GUI agents provide universal interaction capabilities but suffer from severe efficiency bottlenecks due to the massive spatiotemporal redundancy inherent in high-resolution screenshots and historical trajectories. We identify two critical misalignments in existing compression paradigms: the \textit{temporal mismatch}, where uniform history encoding diverges from the agent's ``fading memory'' attention pattern, and the \textit{spatial topology conflict}, where unstructured pruning compromises the grid integrity required for precise coordinate grounding, inducing spatial hallucinations. To address these challenges, we introduce \textbf{GUIPruner}, a training-free framework tailored for high-resolution GUI navigation. It synergizes \textbf{Temporal-Adaptive Resolution (TAR)}, which eliminates historical redundancy via decay-based resizing, and \textbf{Stratified Structure-aware Pruning (SSP)}, which prioritizes interactive foregrounds and semantic anchors while safeguarding global layout. Extensive evaluations across diverse benchmarks demonstrate that GUIPruner consistently achieves state-of-the-art performance, effectively preventing the collapse observed in large-scale models under high compression. Notably, on Qwen2-VL-2B, our method delivers a $\mathbf{3.4\times}$ reduction in FLOPs and a $\mathbf{3.3\times}$ speedup in vision encoding latency while retaining over \textbf{94\%} of the original performance, enabling real-time, high-precision navigation with minimal resource consumption. 
\end{abstract}


\maketitle

\section{Introduction}
The evolution of Multimodal Large Language Models (MLLMs) has fundamentally reshaped the landscape of Graphical User Interface (GUI) agents. Traditional GUI agents~\cite{intro1,intro2,intro3,intro4,chen2025survey} typically rely on textual representations such as HTML or accessibility trees, which are often plagued by noise, excessive sequence lengths, and the requirement for intrusive system-level permissions~\cite{seeclick,intro5}. In contrast, pure-vision-based GUI agents~\cite{seeclick,intro6,intro7,intro8,intro9,intro10} act by directly analyzing screenshots, effectively mimicking human interaction. This paradigm shift empowers agents with robust universality across diverse applications and platforms.

\begin{figure}[t]
    \centering
    \includegraphics[width=\linewidth]{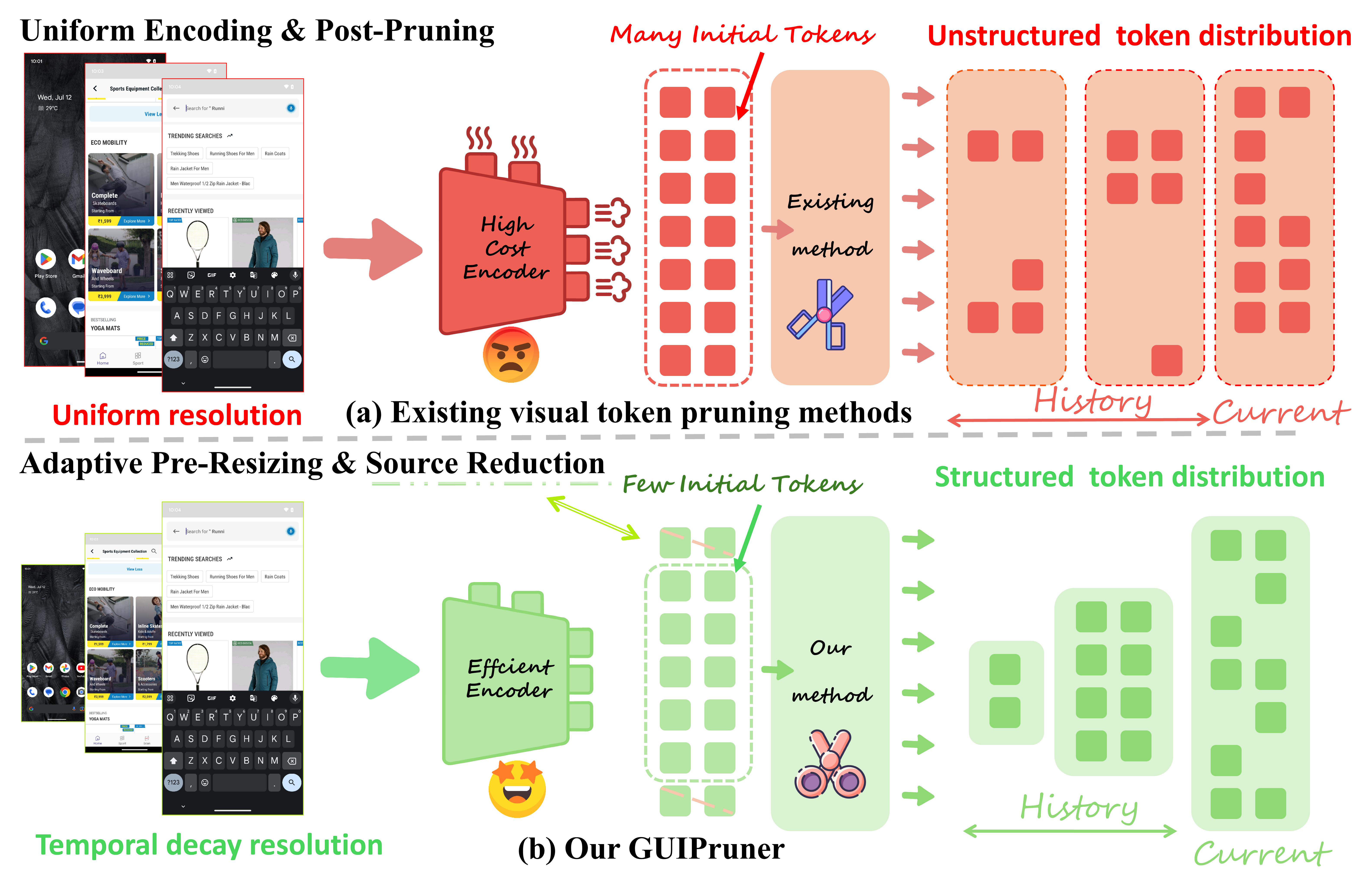}
    \vspace{-0.5cm}

    \caption{\textbf{Paradigm comparison of visual encoding and pruning.} 
    In contrast to conventional pipelines (Top) that incur high redundancy via uniform encoding and disrupt topology via unstructured pruning, \textbf{GUIPruner} (Bottom) implements efficient source-level reduction and structured, topology-preserving compression, ensuring high-precision grounding with minimal token consumption.}
    \label{fig:intro_comparison}
    \vspace{-0.3cm}
\end{figure}

Despite the promising universality of vision-based GUI agents, their efficiency is severely constrained by the spatiotemporal redundancy inherent in contextual modeling. We analyze element and history contexts to uncover two critical challenges that existing methods fail to address.
1) \textbf{The mismatch between high historical resolution and low attention.} As illustrated in Figure~\ref{fig:intro_comparison} (Top), conventional methods typically enforce uniform high-resolution encoding across the entire history, disregarding the temporal dynamics of agent perception. Our analysis, however, reveals a distinct \textbf{``Temporal Decay''} pattern: attention is heavily concentrated on recent frames (the ``Recency Effect'') while distant history receives negligible weight. Consequently, the prevailing ``high resolution, low attention'' configuration incurs massive computational waste, as distant frames necessitate only semantic outlines rather than pixel-level fidelity.
2) \textbf{The conflict between token sparsity and spatial topology.} GUI screenshots are highly sparse, with background tokens often dominating the visual context, accounting for over 60\%. While theoretically compressible, specific background regions (e.g., layout boundaries) serve as essential semantic anchors. Existing general-purpose compression methods~\cite{fastv,mob,cdpruner,divprune} often employ unstructured pruning strategies (Figure~\ref{fig:intro_comparison}, Top Right). Such destructive operations cause two major issues: they disrupt the inherent 2D grid structure required for accurate coordinates grounding, inducing severe \textbf{``spatial hallucinations''}, and they indiscriminately discard structurally meaningful regions, leading to degraded reasoning and potential task failures.

To address these challenges, we propose \textbf{GUIPruner}, a framework tailored for high-resolution GUI navigation. The framework synergizes two modules \textbf{Temporal-Adaptive Resolution (TAR)} and \textbf{Stratified Structure-aware Pruning (SSP)} to efficiently compress historical and element contexts, respectively.
First, to mitigate historical redundancy, we propose \textbf{TAR}, a global-to-local resource scheduling mechanism. Departing from prior works that treat historical frames independently~\cite{guiodyssey,lessismore}, TAR mimics the biological ``fading memory'' by imposing a holistic token budget distributed across the temporal dimension. As depicted in Figure~\ref{fig:intro_comparison} (Bottom Left), we employ a linear decay scheme to allocate token quotas, ensuring resolution monotonically attenuates with temporal distance. Crucially, this pre-computation resizing suppresses token generation at the source, directly curtailing the vision encoder's computational overhead. This mechanism effectively eliminates pixel-level redundancy in distant history while preserving high-fidelity details for the immediate context.
Second, to reconcile the trade-off between token sparsity and spatial topology, we propose \textbf{SSP}, which operates within the shallow layers of the MLLM. Instead of unstructured pruning that disrupts spatial grids (Figure~\ref{fig:intro_comparison}, Top Right), SSP employs a Stratified Budget Allocation Strategy to hierarchically partition the visual context. It prioritizes: (1) preserving high-resolution features of interactive foreground targets (e.g., buttons, input box); (2) retaining semantically salient background regions via attention ranking; and (3) dedicating the residual budget to Uniform Grid Sampling (UGS) (Figure~\ref{fig:intro_comparison}, Bottom Right). UGS acts as a structural safeguard, maintaining a coarse-grained perception of the global layout to prevent ``spatial hallucinations''.

We validate the effectiveness of GUIPruner through rigorous benchmarking against state-of-the-art training-free methods. Evaluated across four diverse datasets (AITW~\cite{aitw}, Mind2Web~\cite{mind2web}, GUI-Odyssey~\cite{guiodyssey}, AndroidControl~\cite{androidcontrol}) and varying model scales (Qwen2-VL-2B~\cite{qwen2} and Qwen2.5-VL-7B~\cite{qwen2_5}), our results confirm its superior robustness. Notably, on the challenging Mind2Web, GUIPruner consistently maintains SOTA performance across varying compression ratios and effectively mitigates the catastrophic collapse observed in Qwen2.5-VL-7B. In terms of efficiency, it achieves a 3.4$\times$ FLOPs reduction and 3.3$\times$ encoder speedup on Qwen2-VL-2B, enabling real-time, high-resolution navigation with minimal resource consumption.

The main contributions are summarized as follows:
\begin{itemize}[leftmargin=*, topsep=0pt, parsep=0pt]
    \item We dissect the spatiotemporal redundancy in GUI agents, identifying ``Temporal Decay'' in history and the ``Sparsity-Topology Conflict'' in the current frame as critical bottlenecks. This analysis uncovers the fundamental limitations of uniform compression paradigms in coordinate-sensitive grounding tasks.
    
    \item We propose \textbf{GUIPruner}, a plug-and-play visual compression framework. By synergizing \textbf{Temporal-Adaptive Resolution (TAR)} and \textbf{Stratified Structure-aware Pruning (SSP)}, our approach dynamically aligns visual encoding with the agent's spatiotemporal cognitive patterns, eliminating redundancy without requiring parameter updates.
    
    \item GUIPruner consistently achieves state-of-the-art performance across diverse benchmarks, effectively mitigating the performance collapse observed in large-scale (7B) models. Empirically, it delivers a \textbf{3.4$\times$ FLOPs reduction} and a \textbf{3.3$\times$ speedup} on Qwen2-VL-2B, enabling efficient real-time deployment.
\end{itemize}

\section{Related Work}
\textbf{GUI Agents.} The development of GUI agents~\cite{related10,related11} has transitioned from pipeline-based systems relying on auxiliary tools (e.g., Accessibility Trees, OCR)~\cite{related1,related2,related3,related4,related12,related13} to end-to-end multimodal models trained on visual corpora. While removing dependencies on inaccessible metadata~\cite{intro7,intro9}, these models incur high computational costs due to the necessity of processing high-resolution inputs and long interaction histories. To mitigate this burden, recent studies have explored various efficiency mechanisms. OdysseyAgent~\cite{guiodyssey} proposes a history resampling module to compress past screenshots, though this introduces additional learnable parameters. Iris~\cite{iris} and ShowUI~\cite{showui} attempt to filter redundant backgrounds using low-level visual cues; however, Iris is restricted to current-frame inputs, while ShowUI suffers from notable performance degradation during inference. Similarly, SimpAgent~\cite{lessismore} and GUI-Rise~\cite{GUI-Rise} employ consistency-guided training or reinforcement learning to summarize historical interactions. In contrast, we propose a training-free visual compression framework explicitly designed to preserve GUI structural integrity, achieving high efficiency without parameter updates.

\textbf{Visual Token Pruning.} 
Multimodal Large Language Models (MLLMs) are crucial for handling complex real-world tasks (e.g., image understanding~\cite{liu2025benchmarking}, code generation~\cite{xiao2025efficientuicoder, xiao2026comuicoder, xiao2024interaction2code, xiao2025designbench, wan2024mrweb, dang2025envisioning}, structural visual design~\cite{tang2025slidecoder}, testing agents~\cite{wan2025automatically}), yet processing high-resolution images imposes a heavy computational burden. While some approaches~\cite{related15-query1,related16-query2,related17-query3,related18-query4,related19-query5} compress tokens using learnable queries (which require extra training), the research focus has shifted toward training-free pruning methods that work directly during inference. 
State-of-the-art methods generally optimize token selection through different strategies: \textit{FastV}~\cite{fastv} filters out tokens with low attention scores in shallow layers. In contrast, \textit{DivPrune}~\cite{divprune} formulates pruning as a Max-Min Diversity Problem (MMDP)~\cite{mmdp} to select the most diverse subset of tokens. Similarly, methods like \textit{MoB}~\cite{mob} and \textit{CDPruner}~\cite{cdpruner} further refine this by aligning tokens with text prompts or using Determinantal Point Processes (DPP)~\cite{dpp}. 
However, these general methods are primarily designed for natural images. They often fail to preserve the dense and structured layouts typical of Graphical User Interfaces (GUIs). To address this, we propose a specialized compression method that effectively captures the unique structural characteristics of GUIs, resolving the limitations of existing approaches.

\section{Preliminary}

\subsection{Problem Formulation}

We formalize the interaction between the GUI agent and the digital environment as an autoregressive trajectory generation process. Given a global user instruction $\mathcal{I}$, the agent operates within a discrete time horizon to autonomously accomplish the task. At each time step $t$, the agent's policy $\pi$ integrates a multimodal input stream consisting of the static instruction $\mathcal{I}$, the current visual state $X_t$, and a retrospective context $\mathcal{H}_t$ to predict the subsequent action $a_t$ (e.g., precise click coordinates). Formally, the history $\mathcal{H}_t$ comprises a sequence of interleaved past observations and actions, denoted as $\mathcal{H}_t = \{(X_{t-k}, a_{t-k})\}_{k=1}^{T}$, where $T$ represents the maximum temporal window size. While this historical trajectory provides essential contextual cues for tracking execution progress, it simultaneously induces a massive accumulation of visual tokens, which significantly impedes efficient real-time inference.

\subsection{Redundancy of Historical Visual Tokens}
\label{subsec:red_his}

During GUI action prediction, incorporating historical information is crucial for the agent to comprehend the current task, particularly within complex graphical interfaces where historical context aids in making robust decisions. Prior research, such as SimpAgent~\cite{lessismore}, indicates that while the current observation is the dominant factor for action prediction, the inclusion of historical observations further improves models' performance. However, this improvement comes at a significant computational cost, as historical context introduces a large number of additional visual tokens, thereby substantially increasing inference overhead.

\begin{figure}[t]
    \centering
    
    \includegraphics[width=\linewidth]{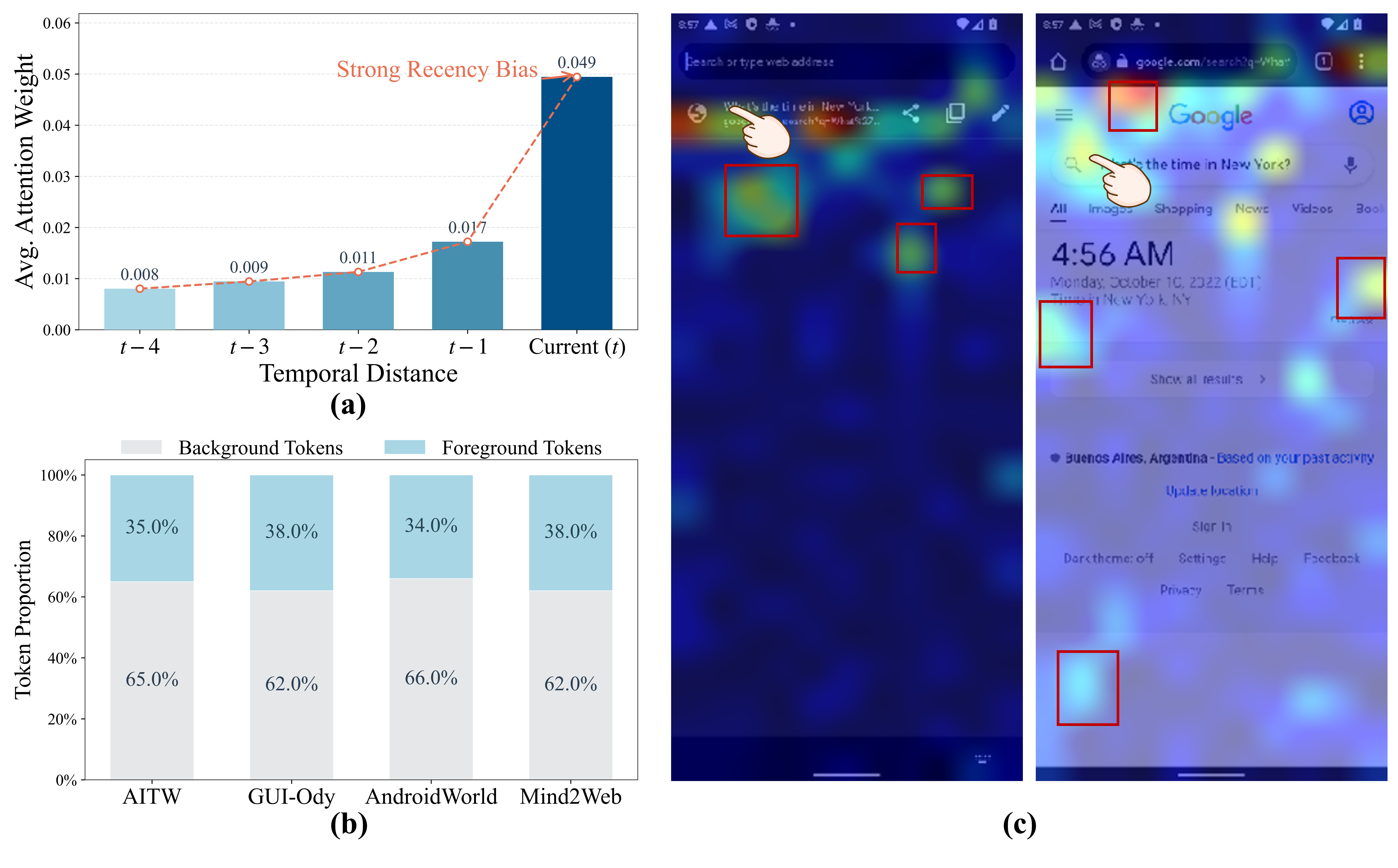}
    
    \vspace{-0.2cm} 
    
    \caption{
    \textbf{(a)} Cross-attention weights exhibit a pronounced \textit{Temporal Decay}, consistent with the ``Recency Effect'' in working memory. 
    \textbf{(b)} Background tokens dominate the visual context across four datasets, indicating significant spatial redundancy. 
    \textbf{(c)} Heatmaps reveal that specific background regions (red boxes) retain high attention as essential semantic anchors, cautioning against indiscriminate pruning.}
    
    \label{fig:preliminary_analysis}

\end{figure}

Historical information functions analogously to task working memory. Human Visual Working Memory (VWM) exhibits a significant ``\textbf{Recency Effect}''~\cite{VWM}, where clear details are retained for recent scenes, while only fuzzy semantic outlines are preserved for distant memories. To investigate whether Multimodal Large Language Models (MLLMs) possess similar characteristics, we visualize the Cross-Attention weights distribution of the agent regarding historical frames during multi-step decision-making processes (as shown in Figure~\ref{fig:preliminary_analysis}(a)).

We observe a pronounced \textbf{Temporal Decay} phenomenon: the agent's attention is heavily concentrated on the most recent $1 \sim 2$ frames to capture the immediate interaction state. Conversely, as the time lag increases, the attention weights assigned to historical frames gradually diminish. However, existing GUI agents overlook this pattern, enforcing a uniform high-resolution encoding for all historical frames regardless of their temporal distance. The ``high resolution, low attention'' configuration results in severe computational waste. This observation inspires our \textbf{Dynamic Resolution Strategy}, which adaptively decays the input resolution as the time lag increases, eliminating pixel-level redundancy while mimicking biological memory mechanisms.

\begin{tcolorbox}[colback=gray!20, colframe=gray!20, width=\columnwidth, left=0.05in, right=0.05in, top=0.05in, bottom=0.05in]
\textbf{Observation 1:} MLLMs' attention over GUI histories decays sharply over time, making uniform high-resolution encoding of distant frames inefficient.
\end{tcolorbox}

\subsection{Redundancy of Current Frame}

To investigate the potential for compressing the current frame, we first conduct a quantitative analysis of visual token composition in high-resolution GUI screenshots. As illustrated in Figure~\ref{fig:preliminary_analysis}(b), due to the inherent sparsity of GUI layouts, background tokens predominantly occupy the visual context (often exceeding 60\%), while foreground tokens representing interactive elements constitute only a minor fraction. This extreme imbalance reveals substantial computational redundancy. However, this does not imply that background tokens can be indiscriminately discarded. By visualizing the attention distribution (as shown in Figure~\ref{fig:preliminary_analysis}(c)), we observe that specific background regions, particularly those layout references, retain high attention scores, suggesting that naive background removal risks stripping away essential semantic context.

\begin{figure*}[t]
    \centering
    
    \includegraphics[width=\linewidth]{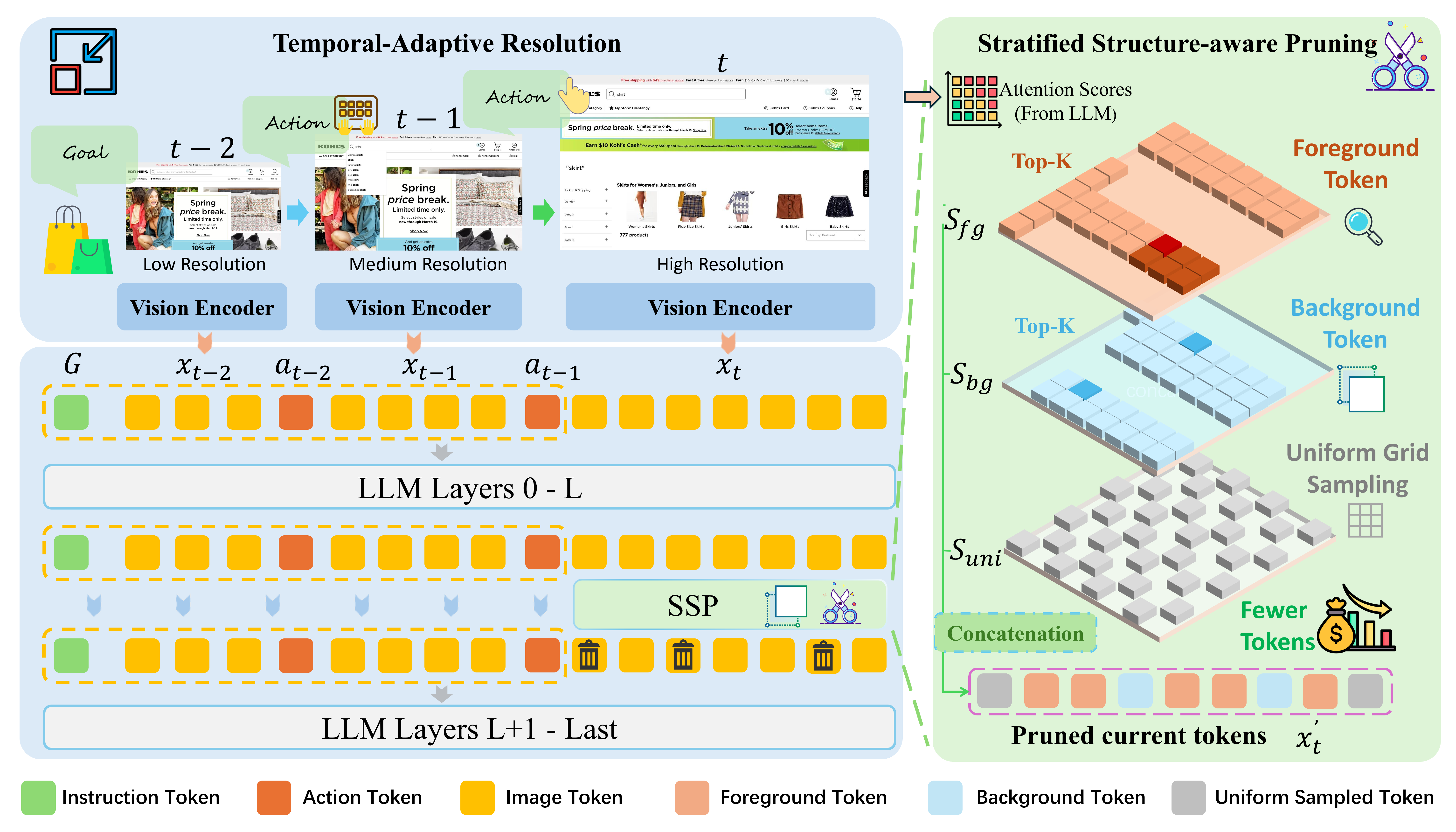}

    \vspace{-0.3cm} 
    
    \caption{\textbf{Overview of the GUIPruner framework.} 
    The framework addresses spatiotemporal redundancy through two synergistic modules: 
    \textbf{(Left) Temporal-Adaptive Resolution (TAR)} mimics biological fading memory by assigning decaying resolution budgets to historical frames based on temporal distance, eliminating redundancy in distant context. 
    \textbf{(Right) Stratified Structure-aware Pruning (SSP)} operates on the current frame within shallow LLM layers. It preserves topological integrity by hierarchically retaining interactive foreground tokens ($S_{fg}$), semantic background anchors ($S_{bg}$), and a uniform structural grid ($S_{uni}$), effectively compressing visual tokens without inducing spatial hallucinations.}
    \label{fig:framework}

\end{figure*}

Crucially, we must address the stringent spatial requirements of GUI tasks. Unlike general visual question answering, GUI agents typically need to output precise pixel coordinates $(x, y)$, which heavily dependents on the integrity of 2D spatial positional encodings. Existing research~\cite{showui} indicates that unstructured token pruning disrupts the original grid integrity of the image. This topological degradation impairs the agent's perception of relative element positions, thereby inducing ``spatial hallucinations'' in coordinate prediction. Consequently, an effective compression strategy must satisfy two requirements: it should drastically reduce background redundancy while strictly preserving the topological structure of the spatial grid to ensure accurate grounding.

\begin{tcolorbox}[colback=gray!20, colframe=gray!20, width=\columnwidth, left=0.05in, right=0.05in, top=0.05in, bottom=0.05in]
\textbf{Observation 2:} Current GUI frames exhibit extreme background redundancy, yet naive token pruning risks disrupting spatial structure and inducing ``spatial hallucinations,'' highlighting the need for topology-preserving compression.
\end{tcolorbox}

\section{Methodology}
\subsection{Overview}
In this section, we detail our proposed framework, GUIPruner, as shown in Figure~\ref{fig:framework}. We introduce two novel mechanisms: \textbf{Temporal-Adaptive Resolution (TAR, \S \ref{subsec:TAR})} for eliminating pixel-level redundancy in historical contexts, and \textbf{Stratified Structure-aware Pruning (SSP, \S \ref{subsec:SSP})} for efficient, topology-preserving encoding of the current interactive state.

\subsection{Temporal-Adaptive Resolution}
\label{subsec:TAR}

To address the temporal redundancy identified in Section~\ref{subsec:red_his}, we propose the \textbf{Temporal-Adaptive Resolution (TAR)} mechanism to dynamically allocate resolution for historical frames. Unlike prior works that treat history frames independently, our method adopts a global-to-local resource scheduling mechanism. It first imposes a holistic constraint on the total computational budget and then distributes this budget across the temporal dimension, mimicking the biological ``fading memory'' where information density decays as time goes by.

\textbf{Global Budget Formulation.} Formally, let the history context at time step $t$ be denoted as $\mathcal{H}_t = \{X_{t-1}, X_{t-2}, \dots, X_{t-T}\}$, where $X_{t-k}$ represents the frame with a temporal lag $k$. To strictly control the computational overhead, we introduce a global hyperparameter $\lambda \in (0, 1]$, termed the History Token Retention Ratio. Given the standard token count per frame $N_{orig} = \frac{HW}{P^2}$ (where $P$ is the patch size, assuming uniform resolution $H \times W$), the total allowable token budget $N_{budget}$ for the entire history sequence is constrained by:
\begin{equation}
    N_{budget} = \lfloor T \times N_{orig} \times \lambda \rfloor,
\end{equation}
which ensures that the agent's memory footprint is bounded by a user-defined ratio $\lambda$, regardless of the history length $T$.

\textbf{Temporal-decay Token Allocation.} We distribute the global budget $N_{budget}$ via a Linear Decay Schedule governed by a retention factor $\gamma \in (0, 1]$. Formally, we define the unnormalized importance weight $w_k$ (interpolating from $1$ down to $\gamma$) and the final allocated token quota $n_k$ as:
\begin{equation}
    w_k = \gamma + (1 - \gamma) \frac{T-k}{T-1}, \quad 
    n_k = N_{budget} \cdot \frac{w_k}{\sum_{j=1}^{T} w_j}.
\end{equation}
This strategy smoothly attenuates resolution for distant frames, allocating higher fidelity to recent observations while preserving minimal context via $\gamma$.

\textbf{Resolution Mapping.} To strictly enforce the allocated token budget $n_k$, we map the discrete quota back to the continuous spatial domain. Given that the token count in ViTs~\cite{vit} scales quadratically with image resolution, we derive the scaling factor $s_k$ for frame $X_{t-k}$ as:
\begin{equation}
    s_k = \sqrt{\frac{n_k}{N_{orig}}}.
\end{equation}
Accordingly, the frame $X_{t-k}$ is resized to $(s_k H, s_k W)$ via bilinear interpolation. This simple projection ensures the actual computational cost aligns with our schedule without introducing any learnable parameters.

\subsection{Stratified Structure-aware Pruning}
\label{subsec:SSP}

To reconcile the efficiency bottleneck of high-resolution inputs with the strict topological requirements of GUI grounding, we propose \textbf{Stratified Structure-aware Pruning (SSP)}. Implemented within the shallow layers of the MLLM, SSP serves as an early-stage structural sieve that refines visual representations before they propagate to deeper reasoning blocks. Unlike prior methods~\cite{fastv,cdpruner,divprune,mob} relying solely on attention scores (which degrade spatial grids) or static priors (which neglect semantic context), SSP dynamically balances foreground regions containing interactive elements, background regions with semantic salience, and global topology under a strict token budget.

\textbf{Importance Estimation and Partitioning.} Given the visual token sequence $\mathcal{T}$ derived from the current frame, we first partition it into a foreground set $\mathcal{T}_{fg}$ and a background set $\mathcal{T}_{bg}$ via edge detection. We then derive an importance score $s_i$ for each token $t_i$ using aggregated attention weights from the shallow transformer layers. By coupling explicit structural priors with implicit attention-based saliency, this step ensures that the pruning policy is conditioned on both the precise geometric boundaries of interactive elements and their semantic relevance.

\textbf{Stratified Retention Policy.} We introduce the Current Token Retention Ratio $\mu \in (0, 1]$ to impose a global constraint on token usage. The total token budget is defined as $K_{total} = \lfloor |\mathcal{T}| \times \mu \rfloor$. To optimize the trade-off between compression and structural integrity, we allocate this budget across three hierarchical levels:

\textbf{(1) Foreground Salience Preservation.} Since foreground regions typically encapsulate direct interaction targets, we prioritize the highest-attention tokens within $\mathcal{T}_{fg}$. We rank tokens by their score $s_i$ and retain the top $\mu$ fraction:
\begin{equation}
    \mathcal{S}_{fg} = \text{TopK}(\mathcal{T}_{fg}, \lfloor |\mathcal{T}_{fg}| \times \mu \rfloor).
\end{equation}
This ensures that high-resolution features of critical interactive components are fully preserved.

\textbf{(2) Background Semantic Retention.} To retain essential contextual cues, we introduce a background saliency factor $\rho \in (0, 1]$. We select the top $\mu \cdot \rho$ fraction of background tokens $\mathcal{T}_{bg}$ based on attention ranking:
\begin{equation}
    \mathcal{S}_{bg} = \text{TopK}(\mathcal{T}_{bg}, \lfloor |\mathcal{T}_{bg}| \times (\mu \cdot \rho) \rfloor).
\end{equation}
This preserves semantic contexts vital for reasoning, even if they are not direct interaction targets.

\textbf{(3) Topological Structure Completion.} To avert the ``spatial collapse'' caused by unstructured pruning, we allocate the residual budget $K_{res} = K_{total} - (|\mathcal{S}_{fg}| + |\mathcal{S}_{bg}|)$ to safeguard global structure. We apply Uniform Grid Sampling (UGS) on the remaining tokens $\mathcal{T}_{remain} = \mathcal{T} \setminus (\mathcal{S}_{fg} \cup \mathcal{S}_{bg})$:
\begin{equation}
    \mathcal{S}_{uni} = \text{UGS}(\mathcal{T}_{remain}, K_{res}).
\end{equation}
This mechanism maintains a coarse-grained representation of the global layout and relative positioning, effectively preventing spatial hallucinations during coordinate grounding.

\textbf{Outcome.} The final compressed sequence $\mathcal{T}_{final} = \mathcal{S}_{fg} \cup \mathcal{S}_{bg} \cup \mathcal{S}_{uni}$ reduces computational overhead in deep layers while rigorously preserving the 2D spatial integrity required for precise navigation. To intuitively illustrate this mechanism, we provide a detailed visualization of the SSP process in Appendix~\ref{app:ssp_vis}.

\section{Experiments}

\subsection{Experimental Setup}

\textbf{Base Models.} To evaluate the scalability and effectiveness of our proposed framework, we employ Qwen2-VL-2B~\cite{qwen2} and Qwen2.5-VL-7B~\cite{qwen2_5} as our foundational Multimodal Large Language Models (MLLMs). To align these general-purpose models with the specific input-output modalities of GUI navigation tasks, we perform supervised fine-tuning on the target datasets. 
During inference, we configure the sampling temperature to $0.01$ for Qwen2-VL-2B and $10^{-6}$ for Qwen2.5-VL-7B. Regarding the specific hyperparameters of our method, we set $\gamma = 0.2$ and $\rho = 0.3$. Comprehensive implementation details regarding the fine-tuning process and additional experimental settings are provided in Appendix~\ref{app:imp_details}.

\textbf{Datasets.} We validate the robustness of our method across four diverse benchmarks, all characterized by long-horizon interactions and sequential visual history. These datasets span distinct GUI ecosystems: GUI-Odyssey, AndroidControl, and AITW represent the extensive Mobile domain, while Mind2Web serves as a rigorous benchmark for the Web domain.

\textbf{Comparative Methods.} We benchmark against state-of-the-art training-free baselines spanning diverse pruning criteria and execution stages. We evaluate Pre-LLM methods, specifically DivPrune~\cite{divprune} and CDPruner~\cite{cdpruner}, which reduce tokens prior to model input. For In-LLM methods, we include FastV~\cite{fastv} and MoB~\cite{mob} , which perform pruning during the inference process.

\begin{table*}[t]
  \centering
    \caption{\textbf{Performance benchmarking on GUI navigation tasks.} We evaluate Qwen2-VL-2B and Qwen2.5-VL-7B across three pruning settings. Abbreviations: ``GUI-Ody'' (GUI-Odyssey), ``M2Web'' (Mind2Web), and ``AndCtrl'' (AndroidControl).}
    \vspace{-0.1cm}
  \label{tab:pruning_comparison_compact}

  \setlength{\tabcolsep}{5pt}

  \resizebox{0.88\textwidth}{!}{%
  \begin{tabular}{l cccc cccc}
    \toprule
    \multirow{2}{*}{\textbf{Method}} & \multicolumn{4}{c}{\textbf{Qwen2-VL-2B}} & \multicolumn{4}{c}{\textbf{Qwen2.5-VL-7B}} \\
    \cmidrule(lr){2-5} \cmidrule(lr){6-9}
     & AITW & M2Web & GUI-Ody & AndCtrl & AITW & M2Web & GUI-Ody & AndCtrl \\
    \midrule
    
    \multicolumn{9}{c}{\textit{Upper Bound (Full Tokens)}} \\
    \midrule
    \rowcolor{gray!10} 
    Original Model & 69.5 & 34.5 & 71.8 & 66.9 & 71.2 & 35.2 & 78.4 & 71.6 \\
    \midrule

    \multicolumn{9}{c}{\textit{Setting I: Retain History 40\% / Current 75\%}} \\
    \midrule
    FastV (ECCV'24)       & 66.2 & 33.4 & 67.3 & 66.3 & 67.9 & 34.3 & 73.5 & 70.8 \\
    DivPrune (CVPR'25)    & 65.3 & 22.8 & 67.1 & 65.6 & 23.2 & 7.7 & 13.6 & 29.4 \\
    CDPruner (NeurIPS'25) & 64.3 & 18.3 & 65.1 & 61.8 & 23.0 & 6.8 & 13.8 & 29.2 \\
    MoB (NeurIPS'25)      & 66.5 & 21.0 & 66.8 & 65.0 & 68.8 & 32.7 & 74.0 & 71.1 \\
    \textbf{GUIPruner}         & \textbf{67.5} & \textbf{33.6} & \textbf{68.1} & \textbf{66.4} & \textbf{69.4} & \textbf{34.7} & \textbf{74.4} & \textbf{71.3} \\
    \midrule

    \multicolumn{9}{c}{\textit{Setting II: Retain History 20\% / Current 75\%}} \\
    \midrule
    FastV (ECCV'24)       & 64.2 & 32.4 & 64.9 & 65.4 & 66.6 & 33.6 & 70.3 & 70.0 \\
    DivPrune (CVPR'25)    & 63.7 & 21.8 & 64.0 & 64.4 & 20.3 & 6.3 & 11.6 & 24.4 \\
    CDPruner (NeurIPS'25) & 63.2 & 18.0 & 63.1 & 60.6 & 20.7 & 6.7 & 11.3 & 24.5 \\
    MoB (NeurIPS'25)      & 64.9 & 20.3 & 64.2 & 64.0 & 66.7 & 31.9 & 70.4 & 70.4 \\
    \textbf{GUIPruner}         & \textbf{66.3} & \textbf{32.9} & \textbf{66.3} & \textbf{65.9} & \textbf{67.9} & \textbf{34.1} & \textbf{73.0} & \textbf{70.9} \\
    \midrule

    \multicolumn{9}{c}{\textit{Setting III: Retain History 10\% / Current 75\%}} \\
    \midrule
    FastV (ECCV'24)       & 62.7 & 31.6 & 62.4 & 64.4 & 65.7 & \textbf{33.1} & 67.7 & 69.2 \\
    DivPrune (CVPR'25)    & 63.0 & 21.0 & 61.2 & 63.6 & 19.7 & 6.6 & 11.5 & 24.2 \\
    CDPruner (NeurIPS'25) & 61.7 & 17.1 & 60.2 & 59.7 & 19.6 & 6.2 & 11.4 & 24.1 \\
    MoB (NeurIPS'25)      & 63.8 & 19.0 & 60.8 & 62.8 & 65.1 & 30.6 & 68.1 & 69.6 \\
    \textbf{GUIPruner}         & \textbf{65.3} & \textbf{31.8} & \textbf{63.8} & \textbf{65.6} & \textbf{66.8} & 32.8 & \textbf{70.1} & \textbf{70.0} \\
    \bottomrule
  \end{tabular}%
  }
\end{table*}

\subsection{Research Questions}

\begin{itemize}[leftmargin=*]
    \item \textbf{RQ1 (Performance \& Scalability):} Can GUIPruner effectively compress visual tokens while preserving or even enhancing navigation performance across diverse GUI domains (Mobile, Web) and varying model scales?
    \item \textbf{RQ2 (Efficiency):} What are the tangible computational benefits of GUIPruner about FLOPs reduction, inference acceleration, and GPU memory overhead compared to SOTA pruning baselines?
    \item \textbf{RQ3 (Component Attribution):} How do the individual components:specifically Temporal-Adaptive Resolution (TAR) for history and Stratified Structure-aware Pruning (SSP) for the current frame—contribute to the overall robustness of the system?
    \item \textbf{RQ4 (Hyperparameter Sensitivity):} How do critical hyperparameters (e.g., temporal decay factor $\gamma$, background saliency factor $\rho$, and pruning layer depth $L$) influence the trade-off between compression rate and grounding accuracy?
\end{itemize}

\subsection{Overall Performance and Scalability (RQ1)}

We evaluate our framework on Qwen2-VL-2B and Qwen2.5-VL-7B across four benchmarks under three history retention settings (10\%, 20\%, 40\%, see Table~\ref{tab:pruning_comparison_compact}). Recognizing the critical role of the current observation, we fix its token retention ratio at 75\%. Our results demonstrate consistent superiority over state-of-the-art baselines, exhibiting robust scalability and adaptability.

\textbf{Universal Effectiveness on Lightweight Models.} On Qwen2-VL-2B, GUIPruner achieves comprehensive dominance across all datasets. Notably, on the challenging Mind2Web benchmark (Setting I), our method attains 33.6\% accuracy, significantly outperforming diversity-based baselines like DivPrune (22.8\%) and effectively matching the uncompressed upper bound (34.5\%). This confirms that combining pixel-level history downsampling with structure-aware pruning offers the most robust information retention for MLLMs with limited capacity.

\textbf{Mitigating Distribution Shift on Large-Scale Models.} On Qwen2.5-VL-7B, baselines like DivPrune and CDPruner suffer from catastrophic performance collapse, notably on Mind2Web (plummeting to 7.7\% and 6.8\%). We attribute this failure to their aggressive pre-LLM pruning, which discards tokens prior to input, severing spatiotemporal dependencies established during fine-tuning. In contrast, our approach circumvents this collapse (recovering 34.7\% on Mind2Web) via a decoupled design: history is spatially downsampled to preserve global layout, while the current frame is pruned within shallow layers. This ensures the model initially perceives the complete visual context, maintaining alignment with the fine-tuned distribution before reducing redundancy, thereby safeguarding early-stage feature extraction.

\textbf{Resilience to High-Resolution Sparsity.} Mind2Web, distinguished by high-resolution viewports and sparse interactive elements, exposes the inherent limitations of metric-based pruning. As evidenced in Table~\ref{tab:pruning_comparison_compact}, diversity or coverage-based methods (e.g., CDPruner, MoB) fail to identify effective tokens in sparse grids, yielding performance inferior to attention-based approaches. We incorporate attention priors to capture semantic relevance, yet unlike FastV's unstructured filtering, we rigorously enforce topological grid integrity. This strict constraint proves essential for preventing spatial misalignment during precise coordinate grounding in high-resolution GUIs, where unstructured selection typically fails.

\subsection{Efficiency Analysis (RQ2)}

We benchmark computational overhead against state-of-the-art pruning methods using Qwen2-VL-2B on a single NVIDIA RTX 4090, configured with $N=4$ history frames on AITW. As detailed in Table~\ref{tab:efficiency_main}, with retention rates $\lambda=0.1$ and $\mu=0.75$, our method achieves a \textbf{3.4$\times$ reduction in total FLOPs} (aggregated across vision encoding, prefill, and decoding phases). This theoretical reduction translates into tangible acceleration: we speed up the Vision Encoder and Prefill stages by \textbf{3.3$\times$} and \textbf{1.9$\times$}, respectively, effectively dismantling the visual encoding bottleneck. Additionally, our approach demonstrates exceptional resource efficiency, capping peak GPU memory at just 5.9 GB, thereby establishing a superior efficiency-performance trade-off. The efficiency analysis for Qwen2.5-VL-7B is provided in the Appendix~\ref{app:efficiency_7b}.

\begin{table}[ht]
    \centering
    \caption{Efficiency comparison on Qwen2-VL-2B (AITW dataset, 4 history frames, RTX 4090). We report the number of tokens, FLOPs, Encoder latency, Prefill latency, and GPU memory usage. \textbf{Bold} indicates the best performance.}
    \label{tab:efficiency_main}
    \resizebox{\linewidth}{!}{
    \Large
    \setlength{\tabcolsep}{4pt} 
    \begin{tabular}{lccccc}
        \toprule
        \textbf{Method} & \textbf{\# Token} & \textbf{\makecell{FLOPs \\ (T)}} & \textbf{\makecell{Encoder Time \\ (ms)}} & \textbf{\makecell{Prefill Time \\ (ms)}} & \textbf{\makecell{GPU Memory \\ (MB)}} \\
        \midrule
        \rowcolor{gray!10} 
        Qwen2-VL-2B & 1320 & 11.5 & 87.9 & 47.5 & 8956 \\
        FastV & 310 & 8.7 ($\times$1.3) & 87.9 ($\times$1.0) & 28.6 ($\times$1.6) & 6798 \\
        DivPrune & 310 & 8.5 ($\times$1.4) & 87.9 ($\times$1.0) & 27.5 ($\times$1.7) & 6180 \\
        CDPruner & 310 & 8.5 ($\times$1.4) & 87.9 ($\times$1.0) & 27.4 ($\times$1.7) & 6180 \\
        MoB & 310 & 8.7 ($\times$1.3) & 87.9 ($\times$1.0) & 28.1 ($\times$1.6) & 6734 \\
        \textbf{GUIPruner} & 310 & \textbf{3.4 ($\times$3.4)} & \textbf{26.6 ($\times$3.3)} & \textbf{24.1 ($\times$1.9)} & \textbf{5902} \\
        \bottomrule
    \end{tabular}
    }
\end{table}

\subsection{Ablation Study (RQ3)}

\subsubsection{Decoupled Analysis of TAR and SSP}

\begin{figure*}[ht] 
    \centering
    
    \includegraphics[width=1.0\linewidth]{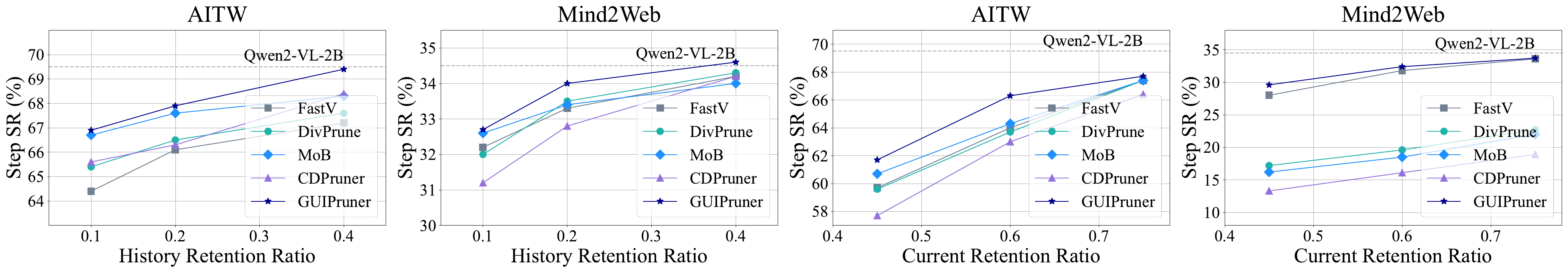}

    \caption{\textbf{Decoupled sensitivity analysis of TAR and SSP on Qwen2-VL-2B.} 
    Left: Evaluation of \textbf{TAR} under varying history retention ratios ($\tau$) on AITW and Mind2Web. 
    Right: Evaluation of \textbf{SSP} under varying current frame retention ratios ($\kappa$). 
    \textbf{GUIPruner} consistently demonstrates superior robustness over baselines, particularly in high-compression regimes.}
    \label{fig:decoupled_analysis}
    
\end{figure*}

To meticulously dissect the individual contributions of our Temporal-Adaptive Resolution (TAR) and Stratified Structure-aware Pruning (SSP) modules, we conduct decoupled experiments using the Qwen2-VL-2B model. The 2B model was selected to isolate the efficacy of our compression strategies from the pre-training distribution shifts observed in larger 7B models. We evaluate our approach on two representative benchmarks: AITW (mobile) and Mind2Web (web). The four line charts in Figure~\ref{fig:decoupled_analysis} visualize the performance trajectories under varying compression intensities.

\textbf{Robustness of TAR.} We evaluated the efficacy of history compression by fixing the current frame retention at 100\% and varying the history retention ratio $\lambda$ across \{10\%, 20\%, 40\%\} (Figure~\ref{fig:decoupled_analysis}, left). Empirical results demonstrate that TAR consistently delivers superior performance across the entire compression spectrum. Notably, at a moderate retention ratio of 40\%, our method achieves near-lossless performance on the AITW benchmark, closely approximating the uncompressed upper bound. More intriguingly, on Mind2Web, TAR even surpasses the original performance of the uncompressed model. This suggests that our ``fading memory'' mechanism not only preserves essential temporal dependencies but also effectively filters out irrelevant historical noise that may otherwise distract the agent.

\textbf{Resilience of SSP.} We assessed current frame compression by retaining the full history context and varying the current frame retention ratio $\mu$ across \{45\%, 60\%, 75\%\} (Figure~\ref{fig:decoupled_analysis}, right). Results demonstrate that SSP consistently outperforms all baselines across compression levels.
On the challenging Mind2Web benchmark, we observe two distinct failure modes in baselines. Metric-based pruners (e.g., MoB, CDPruner, DivPrune) suffer from catastrophic performance collapse, with accuracy languishing below 25\%. While FastV maintains a relatively stable performance curve similar to ours, it consistently lags behind SSP, particularly in the low-retention regime (45\%) where the gap widens. In contrast, SSP achieves the best trade-off, maintaining high accuracy even under aggressive compression. This confirms that explicitly preserving both foreground salience and the global topological grid is essential for precise coordinate grounding in high-resolution GUIs.

\subsubsection{TAR and SSP Deep Dive}

Using Qwen2-VL-2B on AITW and GUI-Odyssey, we validate TAR's adaptive decay against uniform scaling and establish the necessity of SSP's topological grid over random sampling.

\textbf{Effectiveness of Structural Components.} We isolate the contributions of the ``fading memory'' mechanism in TAR and the topological completion in SSP. \textit{(1) Adaptive vs. Uniform History Resolution.} To validate the efficacy of our temporal decay strategy, we benchmark TAR against a Uniform Scaling baseline, which distributes the history budget equidistantly across all past frames. As detailed in Table~\ref{tab:tar_ablation}, our adaptive strategy consistently outperforms the uniform baseline across varying retention ratios ($\lambda \in \{10\%, 20\%, 40\%\}$). This confirms that allocating higher resolution to recent frames effectively aligns with the ``recency bias'' inherent in GUI interactions, whereas uniform allocation inefficiently expends computational resources on distant, stale contexts. \textit{(2) Uniform Grid vs. Random Sampling.} To assess the necessity of topological completion in SSP, we replace the residual Uniform Grid Sampling (UGS) with random sampling while maintaining identical foreground and background retention. Results across current frame retention ratios ($\mu \in \{45\%, 60\%, 75\%\}$) in Table~\ref{tab:ssp_ablation} demonstrate the clear superiority of UGS. This verifies that preserving a deterministic, coarse-grained spatial skeleton is indispensable for preventing grounding hallucinations, whereas unstructured random sampling fails to uphold global layout integrity.

\begin{table}[ht]
    \centering
    \caption{History token allocation ablation. We compare our proposed decay-based strategy (TAR) against a uniform allocation baseline across varying history budgets. "Rel." denotes performance relative to the full-context upper bound.}
    \label{tab:tar_ablation}

    \resizebox{\linewidth}{!}{
        
        \setlength{\tabcolsep}{6pt} 
        \renewcommand{\arraystretch}{1.0} 
        
        \begin{tabular}{p{2.2cm}cccc}
            \toprule
            \textbf{Method} & \textbf{AITW} & \textbf{GUI-Ody} & \textbf{Avg.} & \textbf{Rel.}\\
            \midrule
            \rowcolor{gray!10} 
            Qwen2-VL-2B & \multicolumn{4}{c}{\textit{Upper Bound (Full Tokens)}} \\
           \textcolor{gray}{Vanilla} & \color{gray}69.5 & \color{gray}71.8 & \color{gray}70.6 & \color{gray}100\% \\ 
            \midrule
            
            \rowcolor{gray!10} 
            Qwen2-VL-2B & \multicolumn{4}{c}{\textit{Retain History 40\% / Current 100\%}} \\
            
            TAR w/ Uniform& 68.4 & 70.2 & 69.3  & 98.2\% \\ 
            \textbf{TAR (Ours)} & \textbf{69.4} & \textbf{70.6} & \textbf{70.0} & \textbf{99.2\%}\\
            \midrule
            \rowcolor{gray!10} 
            Qwen2-VL-2B & \multicolumn{4}{c}{\textit{Retain History 20\% / Current 100\%}} \\
            
            TAR w/ Uniform& 67.0 & 67.4 & 67.2  & 95.2\% \\ 
            \textbf{TAR (Ours)} & \textbf{67.9} & \textbf{68.7} & \textbf{68.3} & \textbf{96.7\%}\\
            \midrule
            \rowcolor{gray!10} 
            Qwen2-VL-2B & \multicolumn{4}{c}{\textit{Retain History 10\% / Current 100\%}} \\
            
            TAR w/ Uniform& 66.2 & 64.6 & 65.4  & 92.6\% \\ 
            \textbf{TAR (Ours)} & \textbf{66.9} & \textbf{66.0} & \textbf{66.5} & \textbf{94.2\%} \\
            
            \bottomrule
        \end{tabular}
    } 
\end{table}

\begin{table}[ht]
    \centering
    \caption{Comparison of spatial sampling strategies for current frame pruning. We assess SSP with uniform grid sampling versus random sampling. }
    \label{tab:ssp_ablation}
    
    \resizebox{\linewidth}{!}{ 
        
        \setlength{\tabcolsep}{6pt} 
        \renewcommand{\arraystretch}{1.0} 
        
        \begin{tabular}{p{2.2cm}cccc}
            \toprule
            \textbf{Method} & \textbf{AITW} & \textbf{GUI-Ody} & \textbf{Avg.} & \textbf{Rel.}\\
            \midrule
            \rowcolor{gray!10} 
            Qwen2-VL-2B & \multicolumn{4}{c}{\textit{Upper Bound (Full Tokens)}} \\
\textcolor{gray}{Vanilla} & \color{gray}69.5 & \color{gray}71.8 & \color{gray}70.6 & \color{gray}100\% \\ 
            \midrule
            
            \rowcolor{gray!10} 
            Qwen2-VL-2B & \multicolumn{4}{c}{\textit{Retain History 100\% / Current 75\%}} \\
            
            SSP w/ Random & 66.5 & 69.4 & 68.0  & 96.3\% \\ 
            \textbf{SSP (Ours)} & \textbf{67.7} & \textbf{69.8} & \textbf{68.8} & \textbf{97.4\%}\\
            \midrule
            \rowcolor{gray!10} 
            Qwen2-VL-2B & \multicolumn{4}{c}{\textit{Retain History 100\% / Current 60\%}} \\
            
            SSP w/ Random & 64.9 & 66.1 & 65.5  & 92.8\% \\ 
            \textbf{SSP (Ours)} & \textbf{66.3} & \textbf{66.7} & \textbf{66.5} & \textbf{94.2\%}\\
            \midrule
            \rowcolor{gray!10} 
            Qwen2-VL-2B & \multicolumn{4}{c}{\textit{Retain History 100\% / Current 45\%}} \\
            
            SSP w/ Random & 60.8 & 60.8 & 60.8  & 86.1\% \\ 
            \textbf{SSP (Ours)} & \textbf{61.7} & \textbf{61.4} & \textbf{61.6} & \textbf{87.3\%} \\
            
            \bottomrule
        \end{tabular}

    } 
\end{table}

\subsection{Parameter Study (RQ4)}

\begin{figure}[ht]
    \centering
   
    \includegraphics[width=\linewidth]{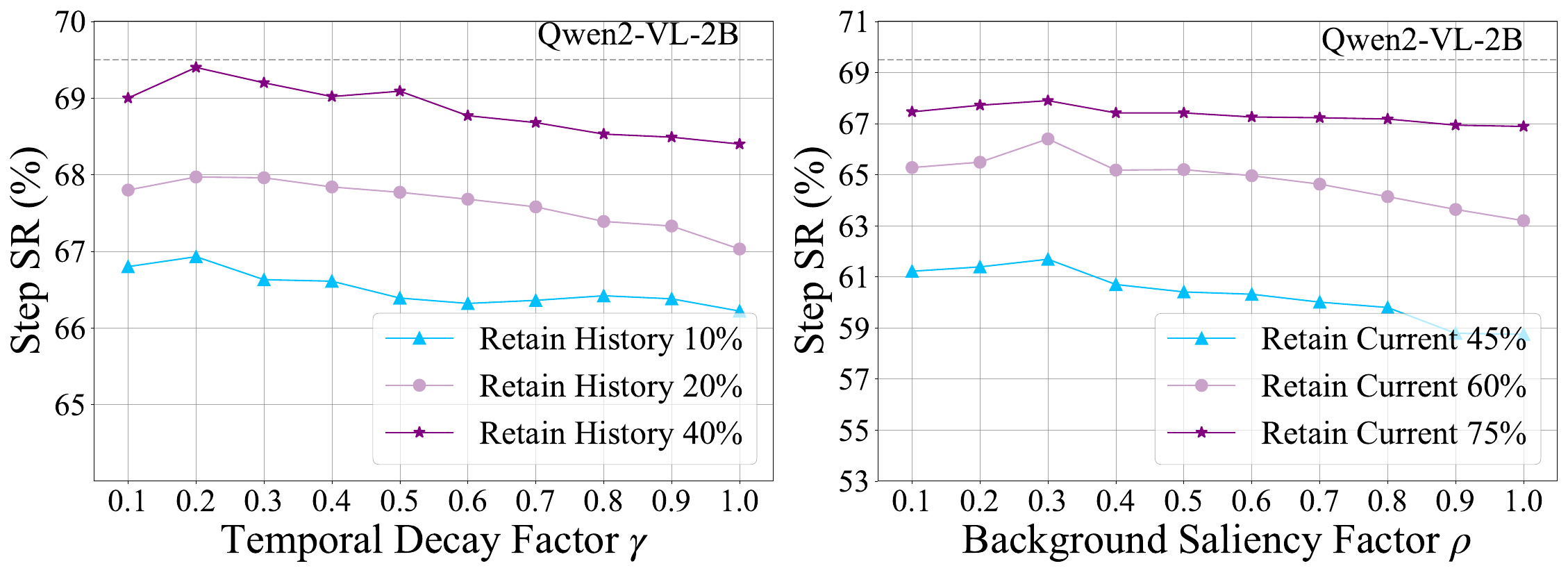}

    \caption{\textbf{Hyperparameter sensitivity analysis on AITW.} We evaluate the Step SR under varying compression intensities: the impact of the temporal decay factor $\gamma$ in TAR, and the impact of the background saliency factor $\rho$ in SSP.}
    \label{fig:hp_ablation}

    \vspace{-0.2cm}
\end{figure}

Using Qwen2-VL-2B on AITW, we analyze the sensitivity of the temporal decay factor $\gamma$, the background saliency factor $\rho$, and the pruning layer depth $L$.

\textbf{Hyperparameter Sensitivity ($\gamma, \rho$).} The results are visualized in Figure~\ref{fig:hp_ablation}, which presents performance curves across three distinct compression rates.

\textit{(1) Temporal Decay Factor $\gamma$.} With current frame retention fixed at 100\%, we vary $\gamma$, the coefficient governing decay steepness ($\gamma=1.0$ signifies uniformity). As shown in Figure~\ref{fig:hp_ablation} (Left), performance peaks at $\gamma = 0.2$ and monotonically degrades as $\gamma \to 1.0$. This confirms that prioritizing immediate history is superior to a uniform distribution, aligning with the agent's reliance on high-fidelity recent context.
\textit{(2) Background Saliency Factor $\rho$.} Conversely, with history frames fully retained to isolate spatial effects, we analyze $\rho$, which modulates the trade-off between local saliency and global structure. Figure~\ref{fig:hp_ablation} (Right) reveals a distinct peak at $\rho = 0.3$. Deviations cause degradation: higher $\rho$ sacrifices layout integrity, while lower $\rho$ captures insufficient semantic context, verifying that a balanced allocation is critical.

\textbf{Pruning Layer Depth ($L$).} With retention ratios fixed at $\lambda=0.1$ and $\mu=0.75$, we systematically investigate the efficiency-performance trade-off by varying the pruning layer insertion depth $L$. As shown in Table~\ref{tab:layer_ablation}, results reveal a distinct non-monotonic trend. 
Pruning at the earliest stage ($L=1$) precipitates a performance collapse (12.4\% Step SR) accompanied by an anomalous surge in FLOPs (3.79T). We attribute this to the immature attention patterns in the first layer, which fail to reliably distinguish interactive semantic regions. This leads to severe grounding hallucinations and output formatting errors, triggering verbose, incorrect generation that paradoxically inflates the total computational cost. 
In contrast, Layer 2 emerges as the optimal operating point, achieving peak performance (65.3\%) while minimizing computational overhead (3.39T). Beyond this point ($L > 2$), performance gradually degrades, and FLOPs linearly increase as additional shallow layers expend resources on the full token sequence before pruning occurs. Qualitative visualizations of attention maps and extended analysis are provided in Appendix~\ref{app:pruning_depth_analysis}.

\begin{table}[h]
    \centering
    \caption{\textbf{Parameter study on the pruning layer depth ($L$) for SSP.} We analyze the trade-off between navigation performance (Step SR) and computational cost (FLOPs) by varying the layer index where pruning is executed.}
    \label{tab:layer_ablation}

    \resizebox{\linewidth}{!}{ 
    
    \setlength{\tabcolsep}{8pt} 
    \renewcommand{\arraystretch}{1.0} 
    
    \begin{tabular}{p{2.2cm}ccccc}
    \toprule 
     \multicolumn{1}{c}{\textbf{$L$}} & \textbf{1} & \textbf{2} &  \textbf{4} &  \textbf{8} &  \textbf{16} \\
    \midrule
    \multicolumn{1}{c}{Step SR} & 12.4 & \textbf{65.3} & 64.4 & 61.1 & 60.6 \\
    \multicolumn{1}{c}{FLOPs (T)} & 3.79 & \textbf{3.39} & 3.45 & 3.48 & 3.53 \\
    \bottomrule
    \end{tabular}
    }
\end{table}

\section{Conclusion}
In this paper, we introduce a novel training-free visual token compression framework tailored for high-resolution GUI navigation agents. Specifically, we decouple spatiotemporal redundancy reduction into two synergistic modules: Temporal-Adaptive Resolution (TAR), which dynamically assigns resolution to capture the ``recency bias'' of interactions, and Stratified Structure-aware Pruning (SSP). SSP employs a stratified budget allocation strategy that prioritizes high-attention foreground and background semantics while rigorously safeguarding the global topological layout through uniform grid sampling. Extensive experiments on representative benchmarks demonstrate that our method achieves state-of-the-art performance across Qwen architectures, effectively preventing the catastrophic performance collapse observed in metric-based baselines on sparse, high-resolution screens. Efficiency analysis further confirms that our approach significantly reduces Vision Encoder latency and FLOPs while maintaining superior grounding accuracy, offering a scalable solution that facilitates the practical deployment of real-time MLLMs in resource-constrained edge applications.

\bibliographystyle{ACM-Reference-Format}
\bibliography{sample-base}

@article{xiao2025efficientuicoder,
  title={EfficientUICoder: Efficient MLLM-based UI Code Generation via Input and Output Token Compression},
  author={Xiao, Jingyu and Zhang, Zhongyi and Wan, Yuxuan and Huo, Yintong and Liu, Yang and Lyu, Michael R},
  journal={arXiv preprint arXiv:2509.12159},
  year={2025}
}

@misc{xiao2026comuicoder,
      title={ComUICoder: Component-based Reusable UI Code Generation for Complex Websites via Semantic Segmentation and Element-wise Feedback}, 
      author={Jingyu Xiao and Jiantong Qin and Shuoqi Li and Man Ho Lam and Yuxuan Wan and Jen-tse Huang and Yintong Huo and Michael R. Lyu},
      year={2026},
      eprint={2602.19276},
      archivePrefix={arXiv},
      primaryClass={cs.SE},
      url={https://arxiv.org/abs/2602.19276}, 
}

@article{xiao2025designbench,
  title={Designbench: A comprehensive benchmark for mllm-based front-end code generation},
  author={Xiao, Jingyu and Wang, Ming and Lam, Man Ho and Wan, Yuxuan and Liu, Junliang and Huo, Yintong and Lyu, Michael R},
  journal={arXiv preprint arXiv:2506.06251},
  year={2025}
}

@article{tang2025slidecoder,
  title={SlideCoder: Layout-aware RAG-enhanced Hierarchical Slide Generation from Design},
  author={Tang, Wenxin and Xiao, Jingyu and Jiang, Wenxuan and Xiao, Xi and Wang, Yuhang and Tang, Xuxin and Li, Qing and Ma, Yuehe and Liu, Junliang and Tang, Shisong and others},
  journal={arXiv preprint arXiv:2506.07964},
  year={2025}
}

@inproceedings{dang2025envisioning,
  title={Envisioning Future Interactive Web Development: Editing Webpage with Natural Language},
  author={Dang, Truong Hai and Xiao, Jingyu and Huo, Yintong},
  booktitle={2025 2nd IEEE/ACM International Conference on AI-powered Software (AIware)},
  pages={61--66},
  year={2025},
  organization={IEEE}
}

@article{liu2025benchmarking,
  title={Benchmarking MLLM-based Web Understanding: Reasoning, Robustness and Safety},
  author={Liu, Junliang and Xiao, Jingyu and Tang, Wenxin and Wang, Wenxuan and Wang, Zhixian and Zhang, Minrui and Yu, Shuanghe},
  journal={arXiv preprint arXiv:2509.21782},
  year={2025}
}

@article{wan2025automatically,
  title={Automatically Generating Web Applications from Requirements Via Multi-Agent Test-Driven Development},
  author={Wan, Yuxuan and Liang, Tingshuo and Xu, Jiakai and Xiao, Jingyu and Huo, Yintong and Lyu, Michael R},
  journal={arXiv preprint arXiv:2509.25297},
  year={2025}
}

@article{wan2024mrweb,
  title={Mrweb: An exploration of generating multi-page resource-aware web code from ui designs},
  author={Wan, Yuxuan and Dong, Yi and Xiao, Jingyu and Huo, Yintong and Wang, Wenxuan and Lyu, Michael R},
  journal={arXiv preprint arXiv:2412.15310},
  year={2024}
}

@article{chen2025survey,
  title={A Survey on the Safety and Security Threats of Computer-Using Agents: JARVIS or Ultron?},
  author={Chen, Ada and Wu, Yongjiang and Zhang, Junyuan and Xiao, Jingyu and Yang, Shu and Huang, Jen-tse and Wang, Kun and Wang, Wenxuan and Wang, Shuai},
  journal={arXiv preprint arXiv:2505.10924},
  year={2025}
}

@article{intro1,
  title={Web agents with world models: Learning and leveraging environment dynamics in web navigation},
  author={Chae, Hyungjoo and Kim, Namyoung and Ong, Kai Tzu-iunn and Gwak, Minju and Song, Gwanwoo and Kim, Jihoon and Kim, Sunghwan and Lee, Dongha and Yeo, Jinyoung},
  journal={arXiv preprint arXiv:2410.13232},
  year={2024}
}

@article{intro2,
  title={Pangu-agent: A fine-tunable generalist agent with structured reasoning},
  author={Christianos, Filippos and Papoudakis, Georgios and Zimmer, Matthieu and Coste, Thomas and Wu, Zhihao and Chen, Jingxuan and Khandelwal, Khyati and Doran, James and Feng, Xidong and Liu, Jiacheng and others},
  journal={arXiv preprint arXiv:2312.14878},
  year={2023}
}

@article{intro3,
  title={Large language models can self-improve at web agent tasks},
  author={Patel, Ajay and Hofmarcher, Markus and Leoveanu-Condrei, Claudiu and Dinu, Marius-Constantin and Callison-Burch, Chris and Hochreiter, Sepp},
  journal={arXiv preprint arXiv:2405.20309},
  year={2024}
}

@article{intro4,
  title={Webrl: Training llm web agents via self-evolving online curriculum reinforcement learning},
  author={Qi, Zehan and Liu, Xiao and Iong, Iat Long and Lai, Hanyu and Sun, Xueqiao and Zhao, Wenyi and Yang, Yu and Yang, Xinyue and Sun, Jiadai and Yao, Shuntian and others},
  journal={arXiv preprint arXiv:2411.02337},
  year={2024}
}

@inproceedings{seeclick,
  title={Seeclick: Harnessing gui grounding for advanced visual gui agents},
  author={Cheng, Kanzhi and Sun, Qiushi and Chu, Yougang and Xu, Fangzhi and YanTao, Li and Zhang, Jianbing and Wu, Zhiyong},
  booktitle={Proceedings of the 62nd Annual Meeting of the Association for Computational Linguistics (Volume 1: Long Papers)},
  pages={9313--9332},
  year={2024}
}

@article{intro5,
  title={Gpt-4v (ision) is a generalist web agent, if grounded},
  author={Zheng, Boyuan and Gou, Boyu and Kil, Jihyung and Sun, Huan and Su, Yu},
  journal={arXiv preprint arXiv:2401.01614},
  year={2024}
}

@article{intro6,
  title={Navigating the digital world as humans do: Universal visual grounding for gui agents},
  author={Gou, Boyu and Wang, Ruohan and Zheng, Boyuan and Xie, Yanan and Chang, Cheng and Shu, Yiheng and Sun, Huan and Su, Yu},
  journal={arXiv preprint arXiv:2410.05243},
  year={2024}
}

@article{intro7,
  title={Mobile-agent-v2: Mobile device operation assistant with effective navigation via multi-agent collaboration},
  author={Wang, Junyang and Xu, Haiyang and Jia, Haitao and Zhang, Xi and Yan, Ming and Shen, Weizhou and Zhang, Ji and Huang, Fei and Sang, Jitao},
  journal={Advances in Neural Information Processing Systems},
  volume={37},
  pages={2686--2710},
  year={2024}
}

@article{intro8,
  title={Mobile-agent: Autonomous multi-modal mobile device agent with visual perception},
  author={Wang, Junyang and Xu, Haiyang and Ye, Jiabo and Yan, Ming and Shen, Weizhou and Zhang, Ji and Huang, Fei and Sang, Jitao},
  journal={arXiv preprint arXiv:2401.16158},
  year={2024}
}

@article{intro9,
  title={Aguvis: Unified pure vision agents for autonomous gui interaction},
  author={Xu, Yiheng and Wang, Zekun and Wang, Junli and Lu, Dunjie and Xie, Tianbao and Saha, Amrita and Sahoo, Doyen and Yu, Tao and Xiong, Caiming},
  journal={arXiv preprint arXiv:2412.04454},
  year={2024}
}

@inproceedings{intro10,
  title={You only look at screens: Multimodal chain-of-action agents},
  author={Zhang, Zhuosheng and Zhang, Aston},
  booktitle={Findings of the Association for Computational Linguistics: ACL 2024},
  pages={3132--3149},
  year={2024}
}

@inproceedings{guiodyssey,
  title={GUIOdyssey: A comprehensive dataset for cross-app GUI navigation on mobile devices},
  author={Lu, Quanfeng and Shao, Wenqi and Liu, Zitao and Du, Lingxiao and Meng, Fanqing and Li, Boxuan and Chen, Botong and Huang, Siyuan and Zhang, Kaipeng and Luo, Ping},
  booktitle={Proceedings of the IEEE/CVF International Conference on Computer Vision},
  pages={22404--22414},
  year={2025}
}

@inproceedings{lessismore,
  title={Less is more: Empowering gui agent with context-aware simplification},
  author={Chen, Gongwei and Zhou, Xurui and Shao, Rui and Lyu, Yibo and Zhou, Kaiwen and Wang, Shuai and Li, Wentao and Li, Yinchuan and Qi, Zhongang and Nie, Liqiang},
  booktitle={Proceedings of the IEEE/CVF International Conference on Computer Vision},
  pages={5901--5911},
  year={2025}
}

@article{aitw,
  title={Androidinthewild: A large-scale dataset for android device control},
  author={Rawles, Christopher and Li, Alice and Rodriguez, Daniel and Riva, Oriana and Lillicrap, Timothy},
  journal={Advances in Neural Information Processing Systems},
  volume={36},
  pages={59708--59728},
  year={2023}
}

@article{mind2web,
  title={Mind2web: Towards a generalist agent for the web},
  author={Deng, Xiang and Gu, Yu and Zheng, Boyuan and Chen, Shijie and Stevens, Sam and Wang, Boshi and Sun, Huan and Su, Yu},
  journal={Advances in Neural Information Processing Systems},
  volume={36},
  pages={28091--28114},
  year={2023}
}

@article{androidcontrol,
  title={On the effects of data scale on computer control agents},
  author={Li, Wei and Bishop, William and Li, Alice and Rawles, Chris and Campbell-Ajala, Folawiyo and Tyamagundlu, Divya and Riva, Oriana},
  journal={arXiv e-prints},
  pages={arXiv--2406},
  year={2024}
}

@inproceedings{fastv,
  title={An image is worth 1/2 tokens after layer 2: Plug-and-play inference acceleration for large vision-language models},
  author={Chen, Liang and Zhao, Haozhe and Liu, Tianyu and Bai, Shuai and Lin, Junyang and Zhou, Chang and Chang, Baobao},
  booktitle={European Conference on Computer Vision},
  pages={19--35},
  year={2024},
  organization={Springer}
}

@inproceedings{divprune,
  title={Divprune: Diversity-based visual token pruning for large multimodal models},
  author={Alvar, Saeed Ranjbar and Singh, Gursimran and Akbari, Mohammad and Zhang, Yong},
  booktitle={Proceedings of the Computer Vision and Pattern Recognition Conference},
  pages={9392--9401},
  year={2025}
}

@article{cdpruner,
  title={Beyond Attention or Similarity: Maximizing Conditional Diversity for Token Pruning in MLLMs},
  author={Zhang, Qizhe and Liu, Mengzhen and Li, Lichen and Lu, Ming and Zhang, Yuan and Pan, Junwen and She, Qi and Zhang, Shanghang},
  journal={arXiv preprint arXiv:2506.10967},
  year={2025}
}

@article{mob,
  title={Why 1+ 1< 1 in Visual Token Pruning: Beyond Naive Integration via Multi-Objective Balanced Covering},
  author={Li, Yangfu and Zhan, Hongjian and Chen, Tianyi and Liu, Qi and Lu, Yue},
  journal={arXiv preprint arXiv:2505.10118},
  year={2025}
}

@inproceedings{iris,
  title={Iris: Breaking gui complexity with adaptive focus and self-refining},
  author={Ge, Zhiqi and Li, Juncheng and Pang, Xinglei and Gao, Minghe and Pan, Kaihang and Lin, Wang and Fei, Hao and Zhang, Wenqiao and Tang, Siliang and Zhuang, Yueting},
  booktitle={Proceedings of the IEEE/CVF International Conference on Computer Vision},
  pages={24559--24568},
  year={2025}
}

@inproceedings{showui,
  title={Showui: One vision-language-action model for gui visual agent},
  author={Lin, Kevin Qinghong and Li, Linjie and Gao, Difei and Yang, Zhengyuan and Wu, Shiwei and Bai, Zechen and Lei, Stan Weixian and Wang, Lijuan and Shou, Mike Zheng},
  booktitle={Proceedings of the Computer Vision and Pattern Recognition Conference},
  pages={19498--19508},
  year={2025}
}

@article{GUI-Rise,
  title={GUI-Rise: Structured Reasoning and History Summarization for GUI Navigation},
  author={Liu, Tao and Wang, Chongyu and Li, Rongjie and Yu, Yingchen and He, Xuming and Song, Bai},
  journal={arXiv preprint arXiv:2510.27210},
  year={2025}
}

@article{mmdp,
  title={GRASP and path relinking for the max--min diversity problem},
  author={Resende, Mauricio GC and Mart{\'\i}, Rafael and Gallego, Micael and Duarte, Abraham},
  journal={Computers \& Operations Research},
  volume={37},
  number={3},
  pages={498--508},
  year={2010},
  publisher={Elsevier}
}

@article{qwen2,
  title={Qwen2-vl: Enhancing vision-language model's perception of the world at any resolution},
  author={Wang, Peng and Bai, Shuai and Tan, Sinan and Wang, Shijie and Fan, Zhihao and Bai, Jinze and Chen, Keqin and Liu, Xuejing and Wang, Jialin and Ge, Wenbin and others},
  journal={arXiv preprint arXiv:2409.12191},
  year={2024}
}

@article{qwen2_5,
  title={Qwen2. 5-vl technical report},
  author={Bai, Shuai and Chen, Keqin and Liu, Xuejing and Wang, Jialin and Ge, Wenbin and Song, Sibo and Dang, Kai and Wang, Peng and Wang, Shijie and Tang, Jun and others},
  journal={arXiv preprint arXiv:2502.13923},
  year={2025}
}

@article{osatlas,
  title={Os-atlas: A foundation action model for generalist gui agents},
  author={Wu, Zhiyong and Wu, Zhenyu and Xu, Fangzhi and Wang, Yian and Sun, Qiushi and Jia, Chengyou and Cheng, Kanzhi and Ding, Zichen and Chen, Liheng and Liang, Paul Pu and others},
  journal={arXiv preprint arXiv:2410.23218},
  year={2024}
}

@article{dpp,
   title={Determinantal Point Processes for Machine Learning},
   volume={5},
   ISSN={1935-8245},
   url={http://dx.doi.org/10.1561/2200000044},
   DOI={10.1561/2200000044},
   number={2–3},
   journal={Foundations and Trends® in Machine Learning},
   publisher={Emerald},
   author={Kulesza, Alex and Taskar, Ben},
   year={2012},
   month=dec, pages={123–286} 
}

@article{clahe,
  title={Realization of the contrast limited adaptive histogram equalization (CLAHE) for real-time image enhancement},
  author={Reza, Ali M},
  journal={Journal of VLSI signal processing systems for signal, image and video technology},
  volume={38},
  number={1},
  pages={35--44},
  year={2004},
  publisher={Springer}
}

@article{canny,
  title={A computational approach to edge detection},
  author={Canny, John},
  journal={IEEE Transactions on pattern analysis and machine intelligence},
  number={6},
  pages={679--698},
  year={2009},
  publisher={Ieee}
}

@article{related1,
  title={Star: Learning diverse robot skill abstractions through rotation-augmented vector quantization},
  author={Li, Hao and Lv, Qi and Shao, Rui and Deng, Xiang and Li, Yinchuan and Hao, Jianye and Nie, Liqiang},
  journal={arXiv preprint arXiv:2506.03863},
  year={2025}
}

@article{related2,
  title={Optimus-1: Hybrid multimodal memory empowered agents excel in long-horizon tasks},
  author={Li, Zaijing and Xie, Yuquan and Shao, Rui and Chen, Gongwei and Jiang, Dongmei and Nie, Liqiang},
  journal={Advances in neural information processing systems},
  volume={37},
  pages={49881--49913},
  year={2024}
}

@article{related3,
  title={Jarvis-1: Open-world multi-task agents with memory-augmented multimodal language models},
  author={Wang, Zihao and Cai, Shaofei and Liu, Anji and Jin, Yonggang and Hou, Jinbing and Zhang, Bowei and Lin, Haowei and He, Zhaofeng and Zheng, Zilong and Yang, Yaodong and others},
  journal={IEEE Transactions on Pattern Analysis and Machine Intelligence},
  year={2024},
  publisher={IEEE}
}

@inproceedings{related4,
  title={React: Synergizing reasoning and acting in language models},
  author={Yao, Shunyu and Zhao, Jeffrey and Yu, Dian and Du, Nan and Shafran, Izhak and Narasimhan, Karthik R and Cao, Yuan},
  booktitle={The eleventh international conference on learning representations},
  year={2022}
}

@inproceedings{related10,
  title={Gui agents: A survey},
  author={Nguyen, Dang and Chen, Jian and Wang, Yu and Wu, Gang and Park, Namyong and Hu, Zhengmian and Lyu, Hanjia and Wu, Junda and Aponte, Ryan and Xia, Yu and others},
  booktitle={Findings of the Association for Computational Linguistics: ACL 2025},
  pages={22522--22538},
  year={2025}
}

@article{related11,
  title={Gui agents with foundation models: A comprehensive survey},
  author={Wang, Shuai and Liu, Weiwen and Chen, Jingxuan and Zhou, Yuqi and Gan, Weinan and Zeng, Xingshan and Che, Yuhan and Yu, Shuai and Hao, Xinlong and Shao, Kun and others},
  journal={arXiv preprint arXiv:2411.04890},
  year={2024}
}

@article{related12,
  title={Agent workflow memory},
  author={Wang, Zora Zhiruo and Mao, Jiayuan and Fried, Daniel and Neubig, Graham},
  journal={arXiv preprint arXiv:2409.07429},
  year={2024}
}

@inproceedings{related13,
  title={Appagent: Multimodal agents as smartphone users},
  author={Zhang, Chi and Yang, Zhao and Liu, Jiaxuan and Li, Yanda and Han, Yucheng and Chen, Xin and Huang, Zebiao and Fu, Bin and Yu, Gang},
  booktitle={Proceedings of the 2025 CHI Conference on Human Factors in Computing Systems},
  pages={1--20},
  year={2025}
}

@article{related15-query1,
  title={Instructblip: Towards general-purpose vision-language models with instruction tuning},
  author={Dai, Wenliang and Li, Junnan and Li, Dongxu and Tiong, Anthony and Zhao, Junqi and Wang, Weisheng and Li, Boyang and Fung, Pascale N and Hoi, Steven},
  journal={Advances in neural information processing systems},
  volume={36},
  pages={49250--49267},
  year={2023}
}

@inproceedings{related16-query2,
  title={Bliva: A simple multimodal llm for better handling of text-rich visual questions},
  author={Hu, Wenbo and Xu, Yifan and Li, Yi and Li, Weiyue and Chen, Zeyuan and Tu, Zhuowen},
  booktitle={Proceedings of the AAAI Conference on Artificial Intelligence},
  volume={38},
  number={3},
  pages={2256--2264},
  year={2024}
}

@inproceedings{related17-query3,
  title={Blip-2: Bootstrapping language-image pre-training with frozen image encoders and large language models},
  author={Li, Junnan and Li, Dongxu and Savarese, Silvio and Hoi, Steven},
  booktitle={International conference on machine learning},
  pages={19730--19742},
  year={2023},
  organization={PMLR}
}

@inproceedings{related18-query4,
  title={Llama-vid: An image is worth 2 tokens in large language models},
  author={Li, Yanwei and Wang, Chengyao and Jia, Jiaya},
  booktitle={European Conference on Computer Vision},
  pages={323--340},
  year={2024},
  organization={Springer}
}

@inproceedings{related19-query5,
  title={Falcon: Resolving visual redundancy and fragmentation in high-resolution multimodal large language models via visual registers},
  author={Zhang, Renshan and Shao, Rui and Chen, Gongwei and Zhang, Miao and Zhou, Kaiwen and Guan, Weili and Nie, Liqiang},
  booktitle={Proceedings of the IEEE/CVF International Conference on Computer Vision},
  pages={23530--23540},
  year={2025}
}

@inproceedings{deepspeed,
  title={Deepspeed: System optimizations enable training deep learning models with over 100 billion parameters},
  author={Rasley, Jeff and Rajbhandari, Samyam and Ruwase, Olatunji and He, Yuxiong},
  booktitle={Proceedings of the 26th ACM SIGKDD international conference on knowledge discovery \& data mining},
  pages={3505--3506},
  year={2020}
}

@article{flashattention,
  title={Flashattention-2: Faster attention with better parallelism and work partitioning},
  author={Dao, Tri},
  journal={arXiv preprint arXiv:2307.08691},
  year={2023}
}

@article{lora,
  title={Lora: Low-rank adaptation of large language models.},
  author={Hu, Edward J and Shen, Yelong and Wallis, Phillip and Allen-Zhu, Zeyuan and Li, Yuanzhi and Wang, Shean and Wang, Lu and Chen, Weizhu and others},
  journal={ICLR},
  volume={1},
  number={2},
  pages={3},
  year={2022}
}

@article {VWM,
	author = {Zokaei, Nahid and Manohar, Sanjay and Husain, Masud and Feredoes, Eva},
	title = {Causal Evidence for a Privileged Working Memory State in Early Visual Cortex},
	volume = {34},
	number = {1},
	pages = {158--162},
	year = {2014},
	doi = {10.1523/JNEUROSCI.2899-13.2014},
	publisher = {Society for Neuroscience},
	issn = {0270-6474},
	URL = {https://www.jneurosci.org/content/34/1/158},
	eprint = {https://www.jneurosci.org/content/34/1/158.full.pdf},
	journal = {Journal of Neuroscience}
}

@article{vit,
  title={A survey on vision transformer},
  author={Han, Kai and Wang, Yunhe and Chen, Hanting and Chen, Xinghao and Guo, Jianyuan and Liu, Zhenhua and Tang, Yehui and Xiao, An and Xu, Chunjing and Xu, Yixing and others},
  journal={IEEE transactions on pattern analysis and machine intelligence},
  volume={45},
  number={1},
  pages={87--110},
  year={2022},
  publisher={IEEE}
}

\appendix
\section*{Appendices}
\section{Details of experimental setup}

\subsection{Model Architecture Details}
We now detail the architectural specifics of the foundation models employed in our framework: Qwen2-VL and Qwen2.5-VL.

\textbf{Qwen2-VL~\cite{qwen2}.} 
Qwen2-VL introduces a Naive Dynamic Resolution mechanism, enabling the processing of images with arbitrary aspect ratios by converting them into variable-length sequences of patches without padding. To capture the spatial structure of these dynamic sequences, it employs Multimodal Rotary Positional Embeddings (M-RoPE), which decompose positional information into temporal, vertical, and horizontal components. Visual features are extracted via a ViT-based encoder and compressed by a factor of four using a $2\times2$ pooling layer before injection into the Qwen2 language backbone.

\textbf{Qwen2.5-VL~\cite{qwen2_5}.} 
Building upon the Qwen2.5 language model, this iteration extends the dynamic processing paradigm to the temporal dimension, supporting native dynamic frame rates for efficient video comprehension. While retaining the M-RoPE mechanism and the C-Former-free architecture of its predecessor, Qwen2.5-VL integrates optimized SwiGLU activations and RMSNorm to ensure numerical stability for long-context multimodal reasoning. The architecture is specifically tuned for fine-grained visual grounding, enabling precise coordinate extraction and structured data parsing.

\subsection{Dataset Details and Preprocessing}

We evaluate our framework across four diverse benchmarks covering mobile and web environments. Below, we detail the specific configurations, splitting protocols, and action spaces for each dataset.

\textbf{Android In The Wild (AITW)~\cite{aitw}.} 
As a large-scale repository for smartphone navigation, AITW contains approximately 30k distinct instructions and 715k execution trajectories. A critical limitation of the original dataset partition is the potential for data leakage, arising from overlapping instructions and highly correlated trajectories across training and test sets. To ensure a rigorous and unbiased evaluation, we strictly adhere to the \textit{instruction-wise split} protocol proposed by SeeClick~\cite{seeclick}, which isolates unique instructions to prevent overfitting. Our data processing pipeline mirrors the settings established in SeeClick. The agent operates within a defined action space comprising 12 primitives: \texttt{Click}, \texttt{Type}, \texttt{Select}, \texttt{Scroll Up/Down/Left/Right}, \texttt{Press Back}, \texttt{Press Home}, \texttt{Press Enter}, \texttt{Status Task Complete}, and \texttt{Status Task Impossible}.

\textbf{GUI-Odyssey~\cite{guiodyssey}.} 
This dataset serves as a comprehensive benchmark for cross-application navigation, featuring 7,735 episodes collected across six distinct mobile devices to test generalization across hardware variations. We align our data preprocessing and evaluation metrics strictly with the original GUI-Odyssey protocols. However, considering the long-horizon nature of this benchmark—characterized by an average of 15 steps per episode~\cite{lessismore}—we conduct our evaluation on a randomly sampled subset comprising one-third of the test set to balance computational efficiency with evaluation rigor. The supported action space consists of 9 distinct operations: \texttt{Click}, \texttt{Scroll}, \texttt{Long Press}, \texttt{Type}, \texttt{Complete}, \texttt{Impossible}, \texttt{Home}, \texttt{Back}, and \texttt{Recent}.

\textbf{AndroidControl~\cite{androidcontrol}.} 
Designed to assess task complexity handling, AndroidControl encompasses 14,548 unique tasks distributed across 833 Android applications. It distinguishes itself by providing both high-level (abstract) and low-level (step-by-step) instructions. For this benchmark, we follow the preprocessing and evaluation standards set by OS-Atlas~\cite{osatlas}. The action space includes 9 primitives: \texttt{Click}, \texttt{Scroll}, \texttt{Long Press}, \texttt{Type}, \texttt{Navigate Home}, \texttt{Navigate Back}, \texttt{Open App}, \texttt{Wait}, and \texttt{Terminate}.

\textbf{Mind2Web~\cite{mind2web}.} 
Mind2Web features over 2,000 tasks across 137 websites, providing raw HTML observations initially tailored for text-based agents.To bridge the modality gap for Vision-Language Models (VLMs), we render these observations into screenshots accompanied by bounding boxes for interactive elements. Since processing full-page screenshots (which can reach resolutions like $1920 \times 12000$) is computationally prohibitive for current VLMs, we adopt the viewport normalization strategy from SeeClick. Specifically, we crop the view around the target element and standardize the resolution to $1920 \times 1080$. The action space for Mind2Web is streamlined to 2 core actions: \texttt{Click} and \texttt{Type}.

\subsection{Comparison methods}
To strictly evaluate the effectiveness of our proposed framework, we compare it against four state-of-the-art, training-free visual token pruning methods. These baselines span distinct pruning paradigms, ranging from attention-based filtering to advanced metric-based optimization.

\textbf{FastV~\cite{fastv}.} 
As a pioneer in attention-guided compression, FastV identifies the ``inefficient attention'' phenomenon where deep layers in MLLMs exhibit sparse attention towards visual tokens. Operating on the hypothesis that tokens receiving low attention scores contribute minimally to semantic reasoning, FastV employs a straightforward filtration strategy. Specifically, it discards visual tokens with the lowest aggregated visual-text attention scores after the second transformer layer. This inference-time pruning mechanism accelerates computation but assumes that attention magnitude is the sole indicator of token importance.

\textbf{DivPrune~\cite{divprune}.} 
Distinct from attention-centric approaches, DivPrune addresses visual redundancy through the lens of geometric diversity. It formulates the token selection process as a Max-Min Diversity Problem (MMDP). Rather than relying on semantic alignment, DivPrune aims to maximize the minimum pairwise distance among the retained tokens. By enforcing spatial separation in the feature space, it seeks to preserve a subset of tokens that are maximally distinct from one another, thereby ensuring broad coverage of the visual field and mitigating the risk of representational collapse.

\textbf{CDPruner~\cite{cdpruner}.} 
This method argues that relying solely on visual similarity or attention scores leads to suboptimal instruction adherence. CDPruner introduces the concept of ``Conditional Diversity'' to bridge this gap. It reformulates the pruning task using a Determinantal Point Process (DPP), a probabilistic framework used to model repulsion between items. By constructing a kernel that conditions visual similarity on the specific textual instruction, CDPruner selects a subset of tokens that maximizes the determinant of the kernel matrix. This approach statistically guarantees that the retained tokens offer high coverage of the image content while remaining strictly relevant to the user's query.

\textbf{MoB~\cite{mob}.} 
MoB provides a rigorous theoretical grounding for token pruning by modeling it as a bi-objective $\epsilon$-covering problem based on the Hausdorff distance. It identifies an intrinsic trade-off between two competing objectives: prompt alignment and visual preservation. To resolve this, MoB derives a closed-form error bound and employs a greedy radius-trading algorithm to optimize the budget allocation dynamically. This method ensures that the selected tokens form a comprehensive cover of the original feature manifold, effectively balancing the reconstruction of visual details with the extraction of instruction-relevant features under strict budget constraints.

\subsection{Implementation Details}
\label{app:imp_details}
\textbf{Hardware Setup.} Fine-tuning was performed on a computational node equipped with four NVIDIA A100 GPUs (80GB), leveraging DeepSpeed ZeRO-2~\cite{deepspeed} and FlashAttention-2~\cite{flashattention} with \texttt{bfloat16} precision. In contrast, all inference and efficiency evaluations were conducted on a single NVIDIA RTX 4090 (24GB) to assess performance on consumer-grade hardware.

\textbf{Base Models.} We employ \textbf{Qwen2-VL-2B} and \textbf{Qwen2.5-VL-7B} as our foundational backbones. These models were selected for their state-of-the-art visual understanding capabilities and dynamic resolution support.

\textbf{Visual Input Configuration.} To balance the trade-off between temporal context and visual resolution under GPU constraints, we follow the visual input configuration set by SimpAgent~\cite{lessismore}. For AITW and GUI-Odyssey, we incorporate 4 historical frames. Furthermore, we apply a global low-resolution setting, where the longest edge of all input images is uniformly rescaled to 512 pixels. Conversely, for AndroidControl and Mind2Web, where discerning fine-grained UI elements is critical, we prioritize visual fidelity over sequence length, retaining 2 high-resolution historical frames.

\textbf{Fine-tuning Strategy.} To align the general-purpose MLLMs with the specific action formatting requirements of each dataset, we performed Supervised Fine-Tuning (SFT) using Low-Rank Adaptation (LoRA)~\cite{lora}. We applied LoRA adapters exclusively to the language model components (linear layers in the attention mechanism) with a rank $r=8$ and a scaling factor $\alpha=16$. Crucially, since our primary objective during this stage is format adaptation rather than injecting new knowledge, we standardized the training duration to \textbf{2 epochs} across all datasets to prevent overfitting and preserve the pre-trained model's generalization abilities.

\subsubsection{Hyperparameter Settings}
We tailored the optimization settings (Batch Size and Learning Rate) for each benchmark to accommodate varying image resolutions and dataset complexities. The configurations on our 4 $\times$ A100 cluster were set as follows:

\begin{itemize}
    \item \textbf{AITW:} We employed a learning rate of $3\text{e-}5$. The per-device batch size was set to 8 with a gradient accumulation step of 1, resulting in a global batch size of 32.
    \item \textbf{Mind2Web:} Given the high resolution of web screenshots, we used a learning rate of $5\text{e-}4$. We reduced the per-device batch size to 2 with a gradient accumulation step of 1, resulting in a global batch size of 8.
    \item \textbf{GUI-Odyssey:} We utilized a learning rate of $3\text{e-}5$. The per-device batch size was set to 4 with 4 gradient accumulation steps, yielding a global effective batch size of 64.
    \item \textbf{AndroidControl:} To handle the task complexity, we employed a higher learning rate of $3\text{e-}4$. We used a per-device batch size of 2 combined with 16 gradient accumulation steps, resulting in a global batch size of 128.
\end{itemize}

\subsubsection{Prompt Construction}
To enable the model to reason about sequential interactions, we organize the input data as an interleaved sequence of historical screenshots and past actions, culminating in the current observation and the natural language instruction. We adopt a structured prompt format where actions are parsed into a JSON-compatible string containing the \texttt{action\_type} and specific arguments.

The specific template for instruction fine-tuning is illustrated in \textbf{Box 1}.

\begin{table}[h]
\centering
\begin{tcolorbox}[
    colback=gray!5!white,
    colframe=gray!75!black,
    title=\textbf{Box 1: Instruction Prompt Template},
    fonttitle=\bfseries,
    boxrule=0.5mm
]
\small
\texttt{"Please generate the next move according to the instruction, previous actions, previous ui screenshot and current ui screenshot.\\
Instruction: What time is it in Paris?.\\
Image\_0: <image>\\
Step\_0: \{\textbackslash"action\_type\textbackslash": Scroll Down\}.\\
Image\_1: <image>\\
Step\_1: \{\textbackslash"action\_type\textbackslash": Click, \textbackslash"click\_point\textbackslash": (545,748)\}.\\
Image\_2: <image>\\
Step\_2: \{\textbackslash"action\_type\textbackslash": Press Home\}.\\
Image\_3: <image>\\
Step\_3: \{\textbackslash"action\_type\textbackslash": Click, \textbackslash"click\_point\textbackslash": (608,669)\}.\\
Image\_4: <image>"}
\end{tcolorbox}
\end{table}

\subsubsection{Edge Detection}

To efficiently categorize visual tokens into foreground ($\mathcal{T}_{fg}$) and background ($\mathcal{T}_{bg}$) sets, we employ a lightweight, training-free detection pipeline tailored for GUI elements. The procedure consists of three stages:

\textbf{1. Preprocessing.} 
The input RGB image is converted to grayscale and smoothed using a Gaussian filter ($5\times5$ kernel) to suppress sensor noise. Crucially, to handle varying contrast levels across different applications, we apply Contrast Limited Adaptive Histogram Equalization (CLAHE)~\cite{clahe} to enhance local boundary definitions before detection.

\textbf{2. Multi-scale Edge Extraction.} 
We utilize the Canny~\cite{canny} edge detector with dual-threshold hysteresis. To ensure robustness, we compute edges at two sensitivity levels and merge the results via a bitwise OR operation. Subsequently, a morphological closing operation (using a $3\times3$ rectangular kernel) is applied to bridge disconnected edge fragments and ensure structural continuity of UI widgets.

\textbf{3. Token Masking.} 
Valid contours are filtered based on area and aspect ratio constraints to eliminate pixel-level artifacts. The resulting binary occupancy mask is then downsampled to the Vision Transformer's patch grid via max-pooling, assigning a token to $\mathcal{T}_{fg}$ if its corresponding patch contains any valid structural edge.

\section{Detailed Algorithms}

In this section, we provide the pseudocode for the two core components of our GUIPruner framework: the \textbf{Temporal-Adaptive Resolution (TAR)} mechanism (Algorithm~\ref{alg:tar}) and the \textbf{Stratified Structure-aware Pruning (SSP)} strategy (Algorithm~\ref{alg:ssp}). These algorithms detail the procedural implementation of the global-to-local resource scheduling and the topology-preserving token selection described in the main text.

\begin{algorithm}[tb]
   \caption{Temporal-Adaptive Resolution (TAR)}
   \label{alg:tar}
\begin{algorithmic}[1]
   \STATE {\bfseries Input:} History context $\mathcal{H}_t = \{X_{t-1}, \dots, X_{t-T}\}$, Original token count $N_{orig}$, History retention ratio $\lambda$, Decay factor $\gamma$.
   \STATE {\bfseries Output:} Resized history frames $\mathcal{H}'_t$.
   
   \STATE \textbf{\textsc{Phase 1}: Calculate Global Budget} 
   \STATE $N_{budget} \leftarrow \lfloor T \times N_{orig} \times \lambda \rfloor$
   \STATE Initialize total weight sum $W_{sum} \leftarrow 0$
   \STATE Initialize empty list for weights $\mathbf{w} \leftarrow []$
   
   \STATE \textbf{\textsc{Phase 2}: Compute Linear Decay Weights}
   \FOR{$k = 1$ {\bfseries to} $T$}
       \STATE $w_k \leftarrow \gamma + (1 - \gamma) \frac{T - k}{T - 1}$ \COMMENT{Higher weight for recent frames ($k \to 1$)}
       \STATE Append $w_k$ to $\mathbf{w}$
       \STATE $W_{sum} \leftarrow W_{sum} + w_k$
   \ENDFOR
   
   \STATE \textbf{\textsc{Phase 3}: Allocate and Resize}
   \STATE $\mathcal{H}'_t \leftarrow \emptyset$
   \FOR{$k = 1$ {\bfseries to} $T$}
       \STATE $n_k \leftarrow N_{budget} \cdot \frac{w_k}{W_{sum}}$ \COMMENT{Token quota for frame $t-k$}
       \STATE $s_k \leftarrow \sqrt{\frac{n_k}{N_{orig}}}$ \COMMENT{Derive spatial scaling factor}
       \STATE $X'_{t-k} \leftarrow \text{Resize}(X_{t-k}, s_k)$ \COMMENT{Bilinear interpolation}
       \STATE Add $X'_{t-k}$ to $\mathcal{H}'_t$
   \ENDFOR
   
   \STATE \bfseries return $\mathcal{H}'_t$
\end{algorithmic}
\end{algorithm}

\begin{algorithm}[tb]
   \caption{Stratified Structure-aware Pruning (SSP)}
   \label{alg:ssp}
\begin{algorithmic}[1]
   \STATE {\bfseries Input:} Visual token sequence $\mathcal{T}$, Global retention ratio $\mu$, Background saliency factor $\rho$.
   \STATE {\bfseries Output:} Compressed token sequence $\mathcal{T}_{final}$.
   
   \STATE \textbf{\textsc{Phase 1}: Partition and Estimation}
   \STATE $\mathcal{T}_{fg}, \mathcal{T}_{bg} \leftarrow \text{EdgeDetectionPartition}(\mathcal{T})$
   \STATE Compute importance scores $\{s_i\}$ for all $t_i \in \mathcal{T}$ using shallow layer attention.
   
   \STATE \textbf{\textsc{Phase 2}: Determine Budget Constraints}
   \STATE $K_{total} \leftarrow \lfloor |\mathcal{T}| \times \mu \rfloor$
   
   \STATE \textbf{\textsc{Phase 3}: Hierarchical Retention}
   \STATE $\triangleright$ \textit{Foreground Salience Preservation} 
   \STATE $K_{fg} \leftarrow \lfloor |\mathcal{T}_{fg}| \times \mu \rfloor$
   \STATE $\mathcal{S}_{fg} \leftarrow \text{TopK}(\mathcal{T}_{fg}, K_{fg}, \text{key}=s_i)$
   
   \STATE $\triangleright$ \textit{Background Semantic Retention}
   \STATE $K_{bg} \leftarrow \lfloor |\mathcal{T}_{bg}| \times (\mu \cdot \rho) \rfloor$
   \STATE $\mathcal{S}_{bg} \leftarrow \text{TopK}(\mathcal{T}_{bg}, K_{bg}, \text{key}=s_i)$
   
   \STATE $\triangleright$ \textit{Topological Structure Completion}
   \STATE $K_{res} \leftarrow K_{total} - (|\mathcal{S}_{fg}| + |\mathcal{S}_{bg}|)$
   \STATE $\mathcal{T}_{remain} \leftarrow \mathcal{T} \setminus (\mathcal{S}_{fg} \cup \mathcal{S}_{bg})$
   \STATE $\mathcal{S}_{uni} \leftarrow \text{UniformGridSampling}(\mathcal{T}_{remain}, K_{res})$ \COMMENT{Safeguard global layout}
   
   \STATE \textbf{\textsc{Phase 4}: Final Aggregation}
   \STATE $\mathcal{T}_{final} \leftarrow \mathcal{S}_{fg} \cup \mathcal{S}_{bg} \cup \mathcal{S}_{uni}$
   
   \STATE \bfseries return $\mathcal{T}_{final}$
\end{algorithmic}
\end{algorithm}

\section{SSP Visualization}
\label{app:ssp_vis}

\begin{figure*}[h]
    \centering
    \includegraphics[width=0.55\linewidth]{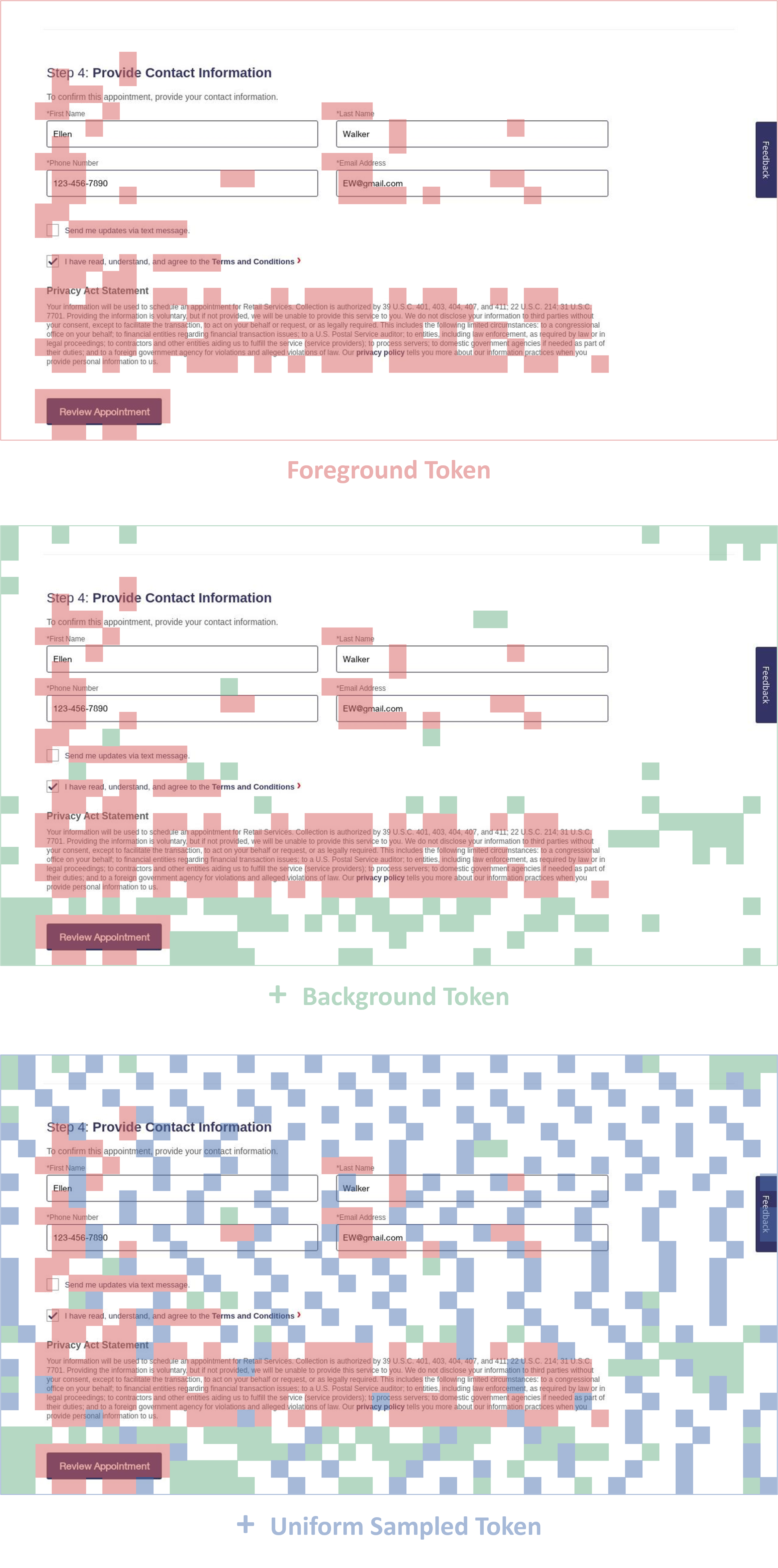}
    \caption{\textbf{Visualizing the SSP Process.} A step-by-step breakdown of token retention. The method progressively layers \textbf{Foreground} (Red), \textbf{Saliency-based Background} (Green), and \textbf{Uniform Grid} (Blue) tokens. This guarantees that both fine-grained interactive elements and the coarse-grained global layout are preserved for the VLM.}
    \label{fig:ssp_vis}
\end{figure*}

Figure~\ref{fig:ssp_vis} illustrates the hierarchical token selection process of our SSP module on a \textbf{Mind2Web} sample, utilizing the \textbf{Qwen2.5-VL-7B} backbone. We configure the current frame retention rate at $\mu=0.45$ and the background saliency factor at $\rho=0.3$. 
The visualization demonstrates the three-stage stratification:
(1) \textbf{Foreground tokens (Red)} representing interactive primitives (e.g., input fields, buttons) are prioritized;
(2) \textbf{Background tokens (Green)} are selected based on saliency scores to capture local context;
(3) \textbf{Uniform Grid tokens (Blue)} fill the remaining budget to preserve the global topological layout.

\section{Analysis of Pruning Depth}
\label{app:pruning_depth_analysis}

\begin{figure*}[h]
    \centering
    \includegraphics[width=\linewidth]{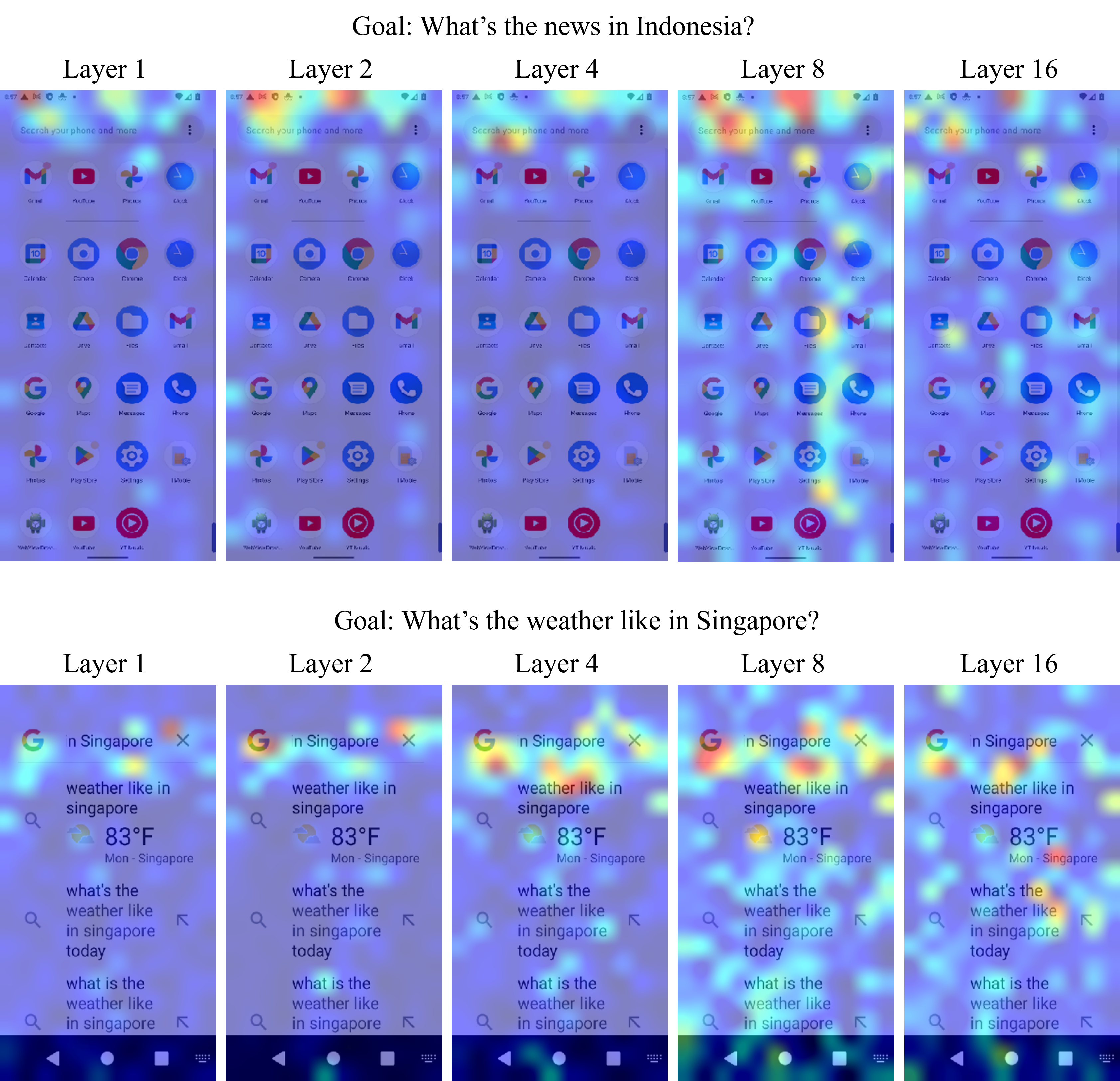} 
    \caption{\textbf{Evolution of LLM Attention Dynamics across Pruning Depths.} 
    The heatmaps illustrate the LLM's visual attention for search-related queries. 
    \textbf{(L1)} Attention is misaligned, failing to cover the target search bar. 
    \textbf{(L2)} Attention is precisely focused on the \textbf{search bar}, providing the most accurate signal for token pruning. 
    \textbf{(L4--L16)} Attention progressively expands to the global context, reducing the selectivity required for efficient pruning.}
    \label{fig:appendix_atten}
\end{figure*}

In the main text, we identified a critical "performance collapse" at pruning depth $L=1$ (12.4\% Step SR) and an optimal operating point at Layer 2 ($L=2$). To elucidate the mechanism governing this non-monotonic trend, we visualize the attention heatmaps of the Large Language Model (LLM) across varying encoder depths ($L \in \{1, 2, 4, 8, 16\}$). Figure~\ref{fig:appendix_atten} illustrates these dynamics for two query-driven scenarios where the interaction target is the search bar.

\textbf{Layer 1: Spatial Misalignment.}
As observed in the first column of Figure~\ref{fig:appendix_atten}, the attention distribution at $L=1$ exhibits severe spatial misalignment. Although the attention is not globally dispersed, it fails to accurately localize the task-relevant region. 
\begin{itemize}
    \item Instead of focusing on the search bar required by the instruction, the attention hotspots (red/yellow regions) land on irrelevant local features or peripheral edges.
\end{itemize}
This phenomenon confirms that at Layer 1, the visual tokens have not yet effectively integrated with the textual instruction to guide localization. Consequently, pruning based on this misaligned map indiscriminately discards the critical search bar tokens while retaining irrelevant noise, directly leading to the performance collapse observed in our experiments.

\textbf{Layer 2: Precise Target Alignment (The Sweet Spot).}
A distinct correction occurs at Layer 2. The attention mechanism successfully fuses the instruction with visual features, correcting the spatial deviation.
\begin{itemize}
    \item As shown in the second column, the attention focus shifts decisively to cover the search bar, aligning perfectly with the interaction target.
\end{itemize}
This precise localization allows the SSP module to retain exactly the tokens representing the interactive component while filtering out the rest. This high "signal-to-noise" ratio in token selection explains why $L=2$ achieves the peak success rate of 65.3\%.

\textbf{Layers 4--16: Contextual Diffusion.}
Contrary to the assumption that deeper layers yield superior pruning masks, our visualizations (Columns 3--5) reveal a trend of contextual diffusion. 
\begin{itemize}
    \item Starting from Layer 4, the attention focus begins to expand from the precise search bar to encompass the broader global context (e.g., the content results below the bar).
\end{itemize}
While this expansion reflects the model's effort to understand the global page semantics, it is detrimental to the specific task of discriminative pruning. By assigning high importance scores to a larger area of the screen, the pruner becomes less selective, retaining redundant background tokens that do not contribute to the immediate interaction. This loss of fine-grained precision, coupled with the increased computational overhead of processing more layers, accounts for the gradual degradation in navigation performance at deeper layers.

\section{Efficiency Analysis}
\label{app:efficiency_7b}
To validate the scalability of our framework, we conducted a systematic benchmarking of computational overhead using Qwen2.5-VL-7B on the AITW benchmark with $N=4$ history frames. As detailed in Table~\ref{tab:efficiency_7b}, with retention rates set to $\lambda=0.1$ and $\mu=0.75$, our method achieves a $3.1\times$ reduction in FLOPs compared to the original model. Crucially, this theoretical gain translates into tangible speedups: we accelerate the Vision Encoder and Prefill stages by $3.2\times$ and $3.0\times$, respectively, effectively dismantling the visual encoding bottleneck that constrains prior baselines. Beyond latency, our approach demonstrates exceptional resource efficiency, capping peak GPU memory usage at just 16.4 GB. Consequently, GUIPruner establishes a superior efficiency-performance trade-off compared to competing methods even on larger-scale backbones.

\begin{table}[t]
    \centering
    \caption{Efficiency comparison on Qwen2.5-VL-7B (AITW dataset, 4 history frames, RTX 4090). We report the number of tokens, FLOPs, Encoder latency, Prefill latency, and GPU memory usage. \textbf{Bold} indicates the best performance.}
    \label{tab:efficiency_7b}
    \resizebox{\linewidth}{!}{
    \Large
    \setlength{\tabcolsep}{4pt} 
    \begin{tabular}{lccccc}
        \toprule
        \textbf{Method} & \textbf{\# Token} & \textbf{\makecell{FLOPs \\ (T)}} & \textbf{\makecell{Encoder Time \\ (ms)}} & \textbf{\makecell{Prefill Time \\ (ms)}} & \textbf{\makecell{GPU Memory \\ (MB)}} \\
        \midrule
        \rowcolor{gray!10} 
        Qwen2.5-VL-7B & 1320 & 27.5 & 308.7 & 156.3 & 18004 \\
        FastV & 310 & 13.8 ($\times$2.0) & 308.7 ($\times$1.0) & 59.1 ($\times$2.6) & 17680 \\
        DivPrune & 310 & 12.7 ($\times$2.2) & 308.7 ($\times$1.0) & 53.6 ($\times$2.9) & 16616 \\
        CDPruner & 310 & 12.7 ($\times$2.2) & 308.7 ($\times$1.0) & 53.8 ($\times$2.9) & 16614 \\
        MoB & 310 & 13.8 ($\times$2.0) & 308.7 ($\times$1.0) & 58.4 ($\times$2.7) & 16768 \\
        \textbf{GUIPruner} & 310 & \textbf{8.9 ($\times$3.1)} & \textbf{97.7 ($\times$3.2)} & \textbf{52.8 ($\times$3.0)} & \textbf{16386} \\
        \bottomrule
    \end{tabular}
    }
\end{table}

\section{Qualitative Analysis}
\label{app:qualitative_vis}

\begin{figure*}[h]
    \centering
    \includegraphics[width=0.7\linewidth]{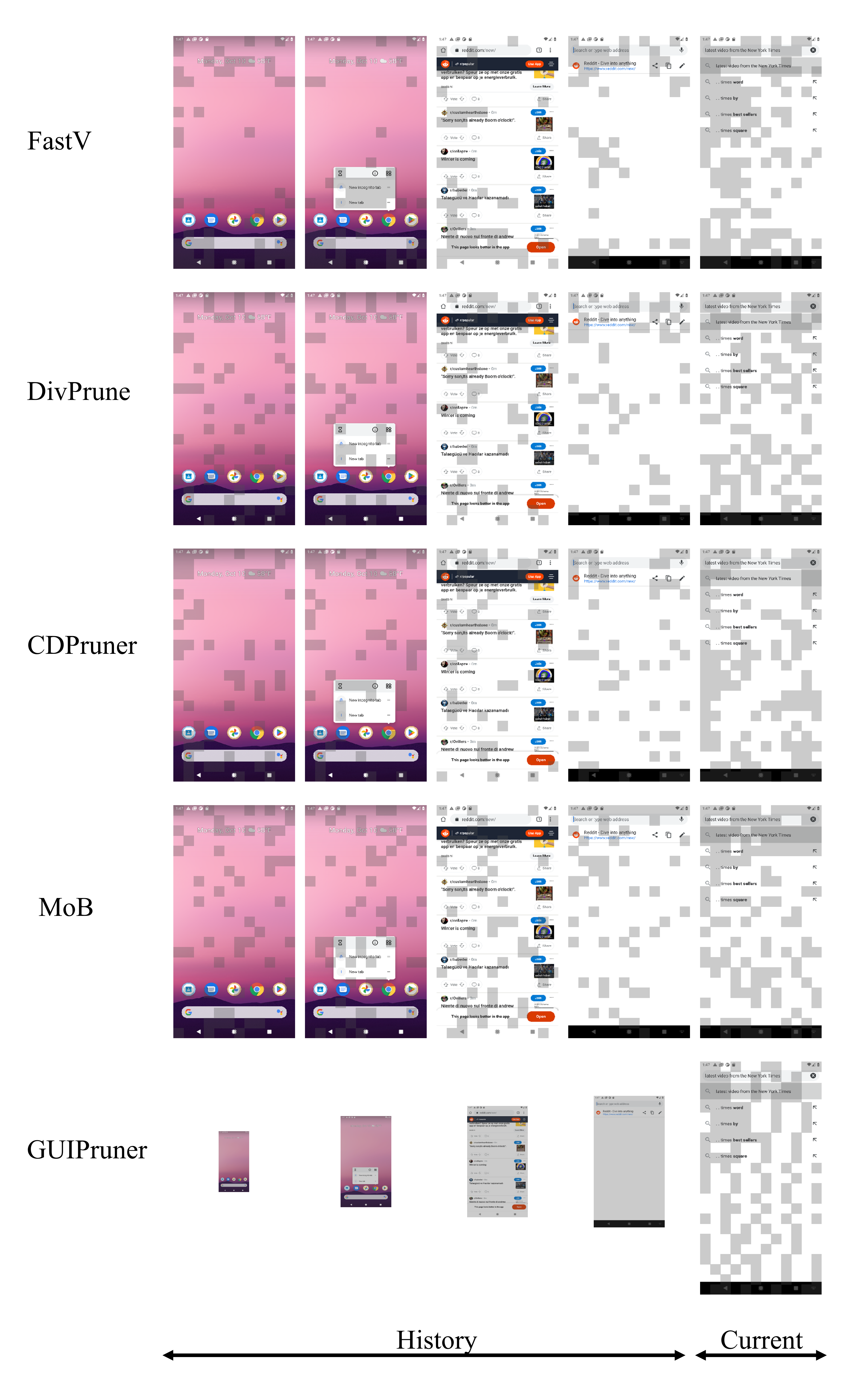}

    \caption{\textbf{Visualization on AITW (Mobile GUI).}  Qualitative comparison demonstrating that baselines  produce fragmented, spatially disjoint patterns. In contrast, GUIPruner preserves the global layout integrity of history frames and enforces a uniform, structure-aware topology for the current frame.}
    \label{fig:vis_aitw}
\end{figure*}

\begin{figure*}[h]
    \centering
    \includegraphics[width=\linewidth]{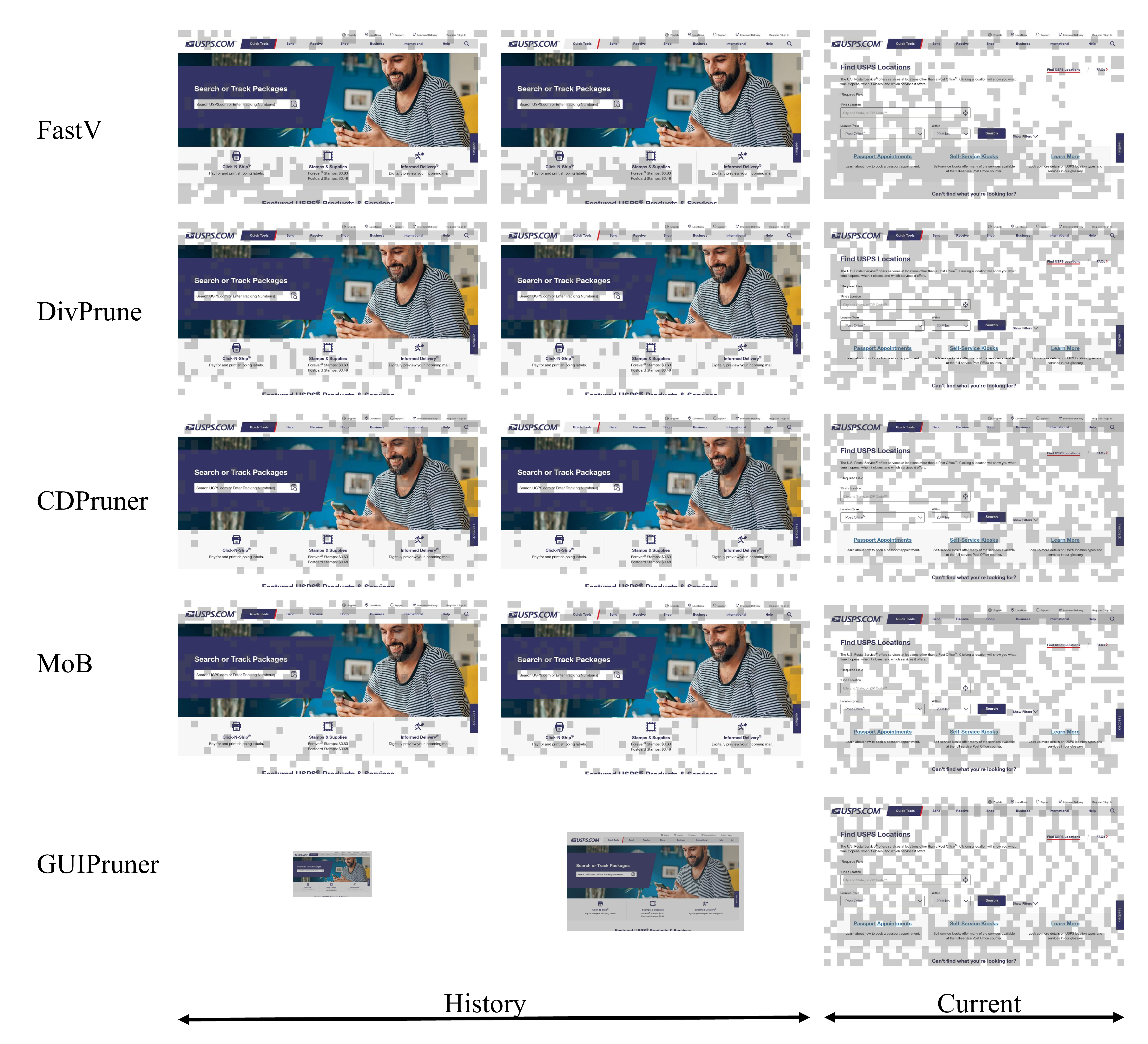}
    \caption{\textbf{Visualization on Mind2Web (Web GUI).}  Qualitative comparison demonstrating that GUIPruner effectively preserves the global layout of historical webpages while precisely highlighting interactive elements in the current frame, mitigating the spatial fragmentation observed in baselines.}
    \label{fig:vis_mind2web}
\end{figure*}

We visualize the retained visual tokens to qualitatively benchmark GUIPruner against state-of-the-art baselines (FastV, DivPrune, CDPruner, and MoB). Experiments are conducted using the Qwen2.5-VL-7B backbone on AITW (Figure~\ref{fig:vis_aitw}) and Mind2Web (Figure~\ref{fig:vis_mind2web}) datasets, with retention rates set to $\lambda=0.2$ (history) and $\mu=0.45$ (current).

\textbf{Structural Integrity in History.} 
For the history columns (left), baseline methods produce highly fragmented representations. This spatial disruption severs the connection between related UI elements (e.g., a button and its label). 
In contrast, GUIPruner preserves global structural integrity. By employing a holistic downsampling strategy rather than sparse token selection, our method maintains the semantic layout. This ensures robust sequential reasoning, even at high compression rates.

\textbf{Balanced Coverage in Current Observations.} 
Current Frame (Right): Unlike baselines that \textbf{over-emphasize} salient regions at the cost of global context, our Stratified Structure-aware Pruning (SSP) balances detailed foregrounds with a structured background. It prioritizes actionable tokens via edge priors while maintaining a uniform grid for the remaining areas. This strategy prevents layout fragmentation, ensuring the vision encoder retains the global spatial integrity essential for precise coordinate grounding.

\section{Limitations}

While GUIPruner establishes a new state-of-the-art in efficient visual navigation, it operates within the constraints of the information bottleneck. As indicated by our ablation analysis, there exists a theoretical lower bound to token reduction; applying pruning at extremely shallow layers or enforcing aggressive compression ratios inevitably risks discarding fine-grained visual cues essential for pixel-level grounding. Consequently, performance degradation is unavoidable when pushing compression beyond the Pareto-optimal frontier. Future work implies exploring more adaptive, instance-aware mechanisms to further optimize this trade-off in highly complex visual scenarios.

\end{document}